\documentclass{ws-ijprai}

\begin{document}

\markboth{D. Soydaner}{A Comparison of Optimization Algorithms for Deep Learning}

%
\catchline{}{}{}{}{}
%

\title{A Comparison of Optimization Algorithms for Deep Learning}

\author{Derya Soydaner
}

\address{Statistics Department, Mimar Sinan Fine Arts University\\
\.{I}stanbul, 34380,Turkey\ \\
\email{derya.soydaner@msgsu.edu.tr}
}

\maketitle


\begin{abstract}
In recent years, we have witnessed the rise of deep learning. Deep neural networks have proved their success in many areas. However, the optimization of these networks has become more difficult as neural networks going deeper and datasets becoming bigger. Therefore, more advanced optimization algorithms have been proposed over the past years. In this study, widely used optimization algorithms for deep learning are examined in detail. To this end, these algorithms called adaptive gradient methods are implemented for both supervised and unsupervised tasks. The behaviour of the algorithms during training and results on four image datasets, namely, MNIST, CIFAR-10, Kaggle Flowers and Labeled Faces in the Wild are compared by pointing out their differences against basic optimization algorithms.
\end{abstract}


\section{Introduction}\label{ch:introduction}

Adaptive gradient methods have been widely used in deep learning. Although one of the most preferred algorithms has been stochastic gradient descent (SGD) for many years, it has difficulties to overcome serious problems such as ill-conditioning and time necessity for large-scale datasets when training deep neural networks. It requires manual tuning of learning rate and difficult to parallelize \cite{le}. Thus, the problems of SGD caused the invention of more advanced algorithms. Nowadays, the optimization algorithms used for deep learning adapt their learning rates during training. Basically, the adaptive gradient methods adjust the learning rate for each parameter. Therefore, when the gradients for some parameters are large, their learning rates are reduced or vice versa.

Until recently, many adaptive methods have been proposed and they become the most commonly used alternatives to SGD. In addition to their high performance on training deep models, another advantage is they are first-order optimization algorithms just as SGD. Thus, they are computationally efficient for training deep neural networks. This work aims to present the most widely used adaptive optimization algorithms that are proven their superiority and compare their working principles. To this end, image processing that is one of the most important areas of deep learning is handled. Firstly, the effects of adaptive gradient methods are observed for image classification task by using convolutional neural networks (CNNs). Secondly, as an unsupervised task, convolutional autoencoders (CAEs) that is one of the quintessential examples of unsupervised learning are used for image reconstruction. Besides, the effects of algorithms are examined by using denoising autoencoders. By this way, the behaviours of the algorithms during training are analyzed in addition to their performances for both supervised and unsupervised learning tasks. 

The rest of the paper is organized as follows. In Section \ref{ch:related_work}, studies about adaptive gradient methods and the most recent variants of them are reviewed briefly. In Section \ref{ch:algorithms}, widely used optimization algorithms in deep learning are explained by showing their update rules and solutions for challenges of training deep networks. SGD and its momentum variants are also mentioned in this section. The experiments and comparative results are in Section \ref{ch:experiments} and conclusion in Section \ref{ch:conclusion}.

\section{Related Work}\label{ch:related_work}

In deep learning literature, working principles and performance analysis of optimization algorithms are widely studied. For example, theoretical guarantees of convergence to criticality for RMSProp and Adam are presented in the setting of optimizing a non-convex objective \cite{soham}. They design experiments to empirically study the convergence and generalization properties of RMSProp and Adam against Nesterov\textsc{\char19}s accelerated gradient method. In another study, conjugate gradient, SGD and limited memory BFGS algorithms are compared \cite{le}. A review is presented on numerical optimization algorithms in the context of machine learning applications \cite{bottouu}. Additionally, similar to this work, an overview of gradient optimization algorithms is summarized \cite{ruder}. 

In this study, most widely used optimization algorihms are examined in the context of deep learning. On the other side, new variants of adaptive methods still have been proposed more recently. For example, new variants of Adam and AMSGrad, called {\it AdaBound} and {\it AMSBound} respectively, are proposed \cite{luo}. They employ dynamic bounds on learning rates to achieve a gradual and smooth transition from adaptive methods to SGD. Also, a new algorithm that adapts the learning rate locally for each parameter separately, and also globally for all parameters together is presented \cite{hayashi}. Another new algorithm, called {\it Nostalgic Adam (NosAdam)}, which places bigger weights on the past gradients than the recent gradients when designing the adaptive learning rate is introduced \cite{huang}. In another study, two variants called {\it SC-Adagrad} and {\it SC-RMSProp} are proposed \cite{mukkamala}. A new adaptive optimization algorithm called {\it YOGI} is presented \cite{zaheer}. It controls the increase in effective learning rate. A novel adaptive learning rate scheme, called {\it ESGD}, based on the equilibration preconditioner is developed \cite{dauphin}. Also, a new algorithm called {\it Adafactor} is presented \cite{shazeer}. Instead of updating parameters scaling by the inverse square roots of exponential moving averages of squared past gradients, Adafactor maintains only the per-row and per-column sums of the moving averages, and estimates the per-parameter second moments based on these sums.

\section{Optimization Algorithms with Adaptive Learning Rates}\label{ch:algorithms}

The choice of the algorithm to optimize a neural network is one of the most important steps. In machine learning, there are three main kinds of optimization methods. First one is called {\it batch} or {\it deterministic} gradient methods that process all training examples simultaneously in a large batch. Second one is called {\it stochastic} or {\it online} methods that use only one example at a time. Today, most algorithms are a blend of the two. During training, they use only a part of training set at each epoch. These algorithms are called {\it minibatch} methods. In deep learning era, minibatch methods are mostly preferred for two major reasons. Firstly, they accelerate the training of neural networks. Secondly, as the minibatches are selected randomly and they are independent, an unbiased estimate of the expected gradient can be computed \cite{goodfellow}.

In this paper, the most widely used minibatch-based adaptive algorithms are examined in detail. Besides, SGD that had been preferred conventionally for a long time is explained alongside its momentum variants, briefly.    

\subsection{Stochastic gradient descent}\label{ch:sgd}

Basically, SGD \cite{robbins} follows the gradient of randomly selected minibatches downhill. In order to train a neural network using SGD, firstly, the gradient estimate is computed by using a loss function. Then, the update at iteration {\it$k$} is applied for parameters {\it$\theta$}. The calculations for each minibatch of {\it$m$} examples from the training set {\it$\big\{{x^{(1)},...,x^{(m)}}\big\}$} with corresponding targets {\it$y^{(i)}$} are as follows:

\begin{equation}
 \hat{g} \leftarrow \frac{1}{m} \nabla_{\theta} \sum \limits_i L(f(x^{(i)};\theta),y^{(i)})
\label{eq:sgd1}
\end{equation}

\begin{equation}
 \theta \leftarrow \theta - \epsilon_{k} \hat{g}
\label{eq:sgd2}
\end{equation}
Here, the learning rate {\it$\epsilon_{k}$} is a very important hyperparameter. The magnitude of the update depends on the learning rate. If it is too large, updates depend too much on recent instances. If it is small, many updates may be need for convergence \cite{alpaydin}. This hyperparameter can be chosen by trial and error. One way is to choose one of the several learning rates that results in the smallest loss function value. This is called {\it line search}. Another way is to monitor the first several epochs and use a learning rate that is higher than the best performing learning rate. In Equation \ref{eq:sgd2}, the learning rate is denoted as {\it$\epsilon_{k}$} at iteration {\it$k$} because in practice, it is necessary to gradually decrease the learning rate over time \cite{goodfellow}. 

\subsubsection{Stochastic gradient descent with momentum}

SGD has difficulty to reach global optimum because of its tendency to oscillate especially on steep surface curves. Noisy or small gradients may also be problematic. The method of momentum \cite{polyak} is designed to accelerate learning in such cases. It aims primarily to solve two problems: Poor conditioning of the Hessian matrix and variance in the stochastic gradient. The idea behind this algorithm is to take a running average by incorporating the previous update in the current change as if there is a momentum due to previous updates \cite{alpaydin}. When SGD is used with momentum, it can converge faster as well as reduced oscillation. 

SGD with momentum uses a variable {\it$v$} called {\it{velocity}}. The velocity is the direction and speed at which the parameters move through parameter space. It is set to an exponentially decaying average of the negative gradient. Also, SGD with momentum requires a new hyperparameter $\alpha$ $\in$ $\big[0,1\big)$ called {\it momentum parameter} that determines how quickly the contributions of previous gradients exponentially decay. The parameters are updated after the velocity update is computed:

\begin{equation}
 v \leftarrow \alpha v - \epsilon \frac{1}{m} \nabla_{\theta} \big(  \sum \limits_{i=1}^{m} L(f(x^{(i)};\theta),y^{(i)}) \big) 
\label{eq:sgdm1}
\end{equation}

\begin{equation}
 \theta \leftarrow \theta + v
\label{eq:sgdm2}
\end{equation}

The velocity {\it v} accumulates the gradient elements. The larger $\alpha$ is relative to $\epsilon$, the more previous gradients affect the current direction. Common values of $\alpha$ used in practice are 0.5, 0.9 and 0.99 \cite{goodfellow}. However, a disadvantage of this algorithm is the requirement of momentum hyperparameter in addition to the learning rate.

\subsubsection{Stochastic gradient descent with Nesterov momentum}

SGD with Nesterov momentum \cite{sutskever} is proposed as a variant of the standard momentum by taking inspiration from Nesterov\textsc{\char19}s accelerated gradient method \cite{nesterov}. The idea is to measure the gradient of the loss function not at the local position but slightly ahead in the direction of the momentum \cite{geron}. This algorithm evaluates  the gradient after the current velocity is applied with Nesterov momentum. Therefore, SGD with Nesterov momentum begins with an interim update for a minibatch \cite{goodfellow}:

\begin{equation}
 \tilde{\theta} \leftarrow \theta + \alpha v
\label{eq:sgdnm1}
\end{equation}

Then, gradient is computed at the interim point. By using this gradient, velocity update is computed. Finally, the parameters are updated: 

\begin{equation}
 g \leftarrow \frac{1}{m} \nabla_{\tilde{\theta}} \sum \limits_{i=1}^{m} L \big(f(x^{(i)};\tilde{\theta}),y^{(i)} \big) 
\label{eq:sgdnm2}
\end{equation}

\begin{equation}
 v \leftarrow \alpha v - \epsilon g 
\label{eq:sgdnm3}
\end{equation}

\begin{equation}
 \theta \leftarrow \theta + v
\label{eq:sgdnm4}
\end{equation}

\subsection{AdaGrad}\label{ch:adagrad}

One of the optimization algorithms that individually adapts the learning rates of model parameters is AdaGrad \cite{duchi}. The parameters with the largest partial derivative of the loss have a rapid decrease in their learning rate, while parameters with small partial derivatives have a relatively small decrease in their learning rate \cite{goodfellow}. This is performed by using all the historical squared values of the gradient. 

AdaGrad uses an additional variable {\it r} for gradient accumulation. In the beginning of this algorithm, the gradient accumulation variable is initialized to zero and gradient is computed for a minibatch: 

\begin{equation}
 g \leftarrow \frac{1}{m} \nabla_{\theta} \sum \limits_i L(f(x^{(i)};\theta),y^{(i)})
\label{eq:adagrad1}
\end{equation}

By using this gradient, the squared gradient is accumulated. Then, the update is computed by scaling learning rates of all parameters inversely proportional to the square root of the sum of all the historical squared values of the gradient. Finally, this update is applied to the model parameters:

\begin{equation}
  r \leftarrow r + g \odot g
\label{eq:adagrad2}
\end{equation}

\begin{equation}
  \Delta\theta \leftarrow - \frac{\epsilon}{\delta + \sqrt{r}} \odot g
\label{eq:adagrad3}
\end{equation}

\begin{equation}
  \theta \leftarrow \theta + \Delta\theta
\label{eq:adagrad4}
\end{equation}
where $\epsilon$ is the global learning rate and $\delta$ is a small constant for numerical stability.

However, AdaGrad has serious disadvantages. Generally, it performs well for simple quadratic problems, but it often stops too early when training neural networks. The learning rate gets scaled down so much that the algorithm ends up stopping entirely before reaching the global optimum \cite{geron}. Also, for training deep neural networks, the accumulation of squared gradients from the beginning of training can result in an excessive decrease in the effective learning rate. AdaGrad performs well for some but not all deep learning models \cite{goodfellow}.

\subsection{AdaDelta}\label{ch:adadelta}

The underlying idea of AdaDelta algorithm is to improve the two main drawbacks of AdaGrad: The continual decay of learning rates throughout training and the need for a manually selected global learning rate. To this end, AdaDelta restricts the window of past gradients to be some fixed size {\it w} instead of accumulating the sum of squared gradients over all time. As mentioned in the previous section, AdaGrad accumulates the squared gradients from each iteration starting at the beginning of training. This accumulated sum continues to grow during training, effectively shrinking the learning rate on each dimension. After many iterations, the learning rate becomes infinitesimally small. With the windowed accumulation AdaGrad becomes a local estimate using recent gradients instead of accumulating to infinity. Thus, learning continues to make progress even after many iterations of updates have been done \cite{zeiler}.

Since storing {\it $w$} previous squared gradients is inefficient, AdaDelta implements this accumulation as an exponentially decaying average of the squared gradients. Assuming this running average is {\it $E\big[g^2\big]_{t}$} at time {\it $t$}, gradient accumulation is computed as follows:   

\begin{equation}
 E\big[g^2\big]_{t} = \rho E\big[g^2\big]_{t-1} + \big(1-\rho\big) g^2_{t} 
\label{eq:adadelta1}
\end{equation}
where {\it $\rho$} is a decay constant similar to that used in the momentum method. Since it is required the square root of this quantity in the parameter updates, this effectively becomes the root mean square (RMS) of previous squared gradients up to time {\it $t$}:

\begin{equation}
 RMS\big[g\big]_{t} = \sqrt{E\big[g^2\big]_{t} + \delta} 
\label{eq:adadelta2}
\end{equation}
where {\it $\delta$} is again a small constant. Based on this RMS, the parameter update is computed, updates are accumulated and the parameters are updated, respectively:

\begin{equation}
 \Delta \theta_{t} = - \frac{RMS\big[\Delta\theta\big]_{t-1}}{RMS\big[g\big]_{t}} g_{t}
\label{eq:adadelta3}
\end{equation}

\begin{equation}
 E\big[\Delta\theta^2\big]_{t} = \rho E\big[\Delta\theta^2\big]_{t-1} + \big(1-\rho\big) \Delta\theta^2_{t}
\label{eq:adadelta4}
\end{equation}

\begin{equation}
 \theta_{t+1} = \theta_{t} + \Delta\theta_{t}
\label{eq:adadelta5}
\end{equation}

The advantage of AdaDelta is that it requires no manual tuning of a learning rate and appears robust to noisy gradient information, different model architectures, various data modalities and selection of hyperparameters \cite{zeiler}. 

\subsection{RMSProp}\label{ch:rmsprop}

Another algorithm that modifies AdaGrad is RMSProp \cite{hinton12}. It is proposed to perform better in the nonconvex setting by changing the gradient accumulation into an exponentially weighted moving average. As mentioned in Section \ref{ch:adagrad}, AdaGrad shrinks the learning rate according to the entire history of the squared gradient. Instead, RMSProp uses an exponentially decaying average to discard history from the extreme past so that it can converge rapidly after finding a convex bowl \cite{goodfellow}.

In order to implement RMSProp, squared gradient is accumulated after computing gradient:

\begin{equation}
  r \leftarrow \rho r + (1-\rho)g \odot g
\label{eq:rms_squared_gradient}
\end{equation}
where {\it $\rho$} is the decay rate. Then parameter update is computed and applied as follows:

\begin{equation}
  \Delta\theta = - \frac{\epsilon}{\sqrt{\delta+r}} \odot g 
\label{eq:rms_update}
\end{equation}

\begin{equation}
  \theta \leftarrow \theta + \Delta\theta
\label{eq:rms_apply_update}
\end{equation}

\subsection{Adam}\label{ch:adam}

Adam is one of the most widely used optimization algorithms in deep learning. The name Adam is derived from adaptive moment estimation because it computes individual adaptive learning rates for different parameters from estimates of first and second moments of the gradients. Adam combines the advantages of AdaGrad which works well with sparse gradients and RMSProp which works well in online and non-stationary settings \cite{kingma}.

There are some important properties of Adam. Firstly, momentum is incorporated directly as an estimate of the first-order moment of the gradient. Also, Adam includes bias corrections to the estimate of both the first-order moments and the second-order moments to account for their initialization at the origin \cite{goodfellow}. The algorithm updates exponential moving averages of the gradient {\it $m_{t}$} and the squared gradient {\it $u_{t}$} where the hyperparameters {\it $\rho_{1}$} ve {\it $\rho_{2}$} control the exponential decay rates of these moving averages. The moving averages themselves are estimates of the first moment (the mean) and the second raw moment (the uncentered variance) of the gradient \cite{kingma}.  

Adam algorithm requires first and second moment variables {\it $m$} and {\it $u$}. After computing gradient, biased first and second moment estimates are updated at time step {\it $t$} respectively:

\begin{equation}
  m_{t} \leftarrow \rho_1 m_{t-1} + (1-\rho_1)g_{t}
\label{eq:first_moment_estimate}
\end{equation}

\begin{equation}
  u_{t} \leftarrow \rho_2 u_{t-1} + (1-\rho_2)g \odot g 
\label{eq:second_moment_estimate}
\end{equation}

Then, bias is corrected in first and second moments. By using the corrected moment estimates parameter updates are calculated and applied:

\begin{equation}
  \hat{m}_{t} \leftarrow \frac{m_{t}}{1-\rho_1^t}
\label{eq:correct_first_bias}
\end{equation}

\begin{equation}
  \hat{u}_{t} \leftarrow \frac{u_{t}}{1-\rho_2^t}
\label{eq:correct_second_bias}
\end{equation}

\begin{equation}
  \Delta \theta = - \epsilon \frac{\hat{m}_{t}}{\sqrt{\hat{u}_{t}}+\delta}
\label{eq:adam_update}
\end{equation}

\begin{equation}
  \theta_{t} \leftarrow \theta_{t-1} + \Delta\theta
\label{eq:adam_apply_update}
\end{equation}

Adam has many advantages. First of all, it requires a little tuning for the learning rate. Also, it is a method that is straightforward to implement and invariant to diagonal rescaling of gradients. It is computationally efficient as well as a little memory requirements. Besides, Adam is appropriate for non-stationary objectives and problems with very noisy and sparse gradients \cite{kingma}.

\subsection{AdaMax}\label{ch:adamax}

AdaMax is proposed as an extension of Adam. It is a variant of Adam based on the infinity norm. In Adam, update rule for individual weights is to scale their gradients inversely proportional to a {\it $L^2$} norm of their individual current and past gradients. AdaMax is based on the idea that {\it $L^2$} norm based update rule can be generalized to a {\it $L^p$} norm based update rule.

AdaMax algorithm begins with calculating gradients w.r.t. stochastic objective at time step {\it $t$}, as usual. Then, biased first moment estimate and exponentially weighted infinity norm are computed. By using them, the model parameters are updated. These steps are defined below, respectively:   

\begin{equation}
  g_{t} \leftarrow \nabla_{\theta} f_{t}\big(\theta_{t-1}\big)
\label{eq:adamax1}
\end{equation}

\begin{equation}
  m_{t} \leftarrow \rho_{1} m_{t-1} + \big(1-\rho_{1}\big) g_{t}
\label{eq:adamax2}
\end{equation}

\begin{equation}
 \gamma_{t} \leftarrow max\big(\rho_{2} \gamma_{t-1}, |g_{t}| \big)
\label{eq:adamax3}
\end{equation}

\begin{equation}
 \theta_{t} \leftarrow \theta_{t-1} - \big(\epsilon / \big(1-\rho^t_{1} \big) \big) m_{t} / \gamma_{t} 
\label{eq:adamax4}
\end{equation}

It is showed that if AdaMax is preferred as optimization algorithm, there is no need to correct for initialization bias. Besides, the magnitude of parameter updates has a simpler bound with AdaMax than Adam \cite{kingma}.

\subsection{Nadam}\label{ch:nadam}

Nadam (Nesterov-accelerated adaptive moment estimation) modifies Adam's momentum component with Nesterov\textsc{\char19}s accelerated gradient. Thus, Nadam aims to improve the speed of convergence and the quality of the learned models \cite{dozat}.

Similar to Adam, after computing gradient, first and second moment variables are updated as in Equations \ref{eq:nadam2} and \ref{eq:nadam3}. Then, corrected moments are computed and parameters are updated as in following equations: 

\begin{equation}
  m_{t} \leftarrow \rho_{t} m_{t-1} + \big(1-\rho_{t}\big) g_{t}
\label{eq:nadam2}
\end{equation}

\begin{equation}
  u_{t} \leftarrow v u_{t-1} + \big(1-v\big) g^2_t
\label{eq:nadam3}
\end{equation}

\begin{equation}
  \hat{m} \leftarrow \big(\rho_{t+1} m_{t} / \big(1-\prod_{i=1}^{t+1} \rho_{i}\big)\big) + \big( \big(1-\rho_{t}\big)g_{t} / \big(1-\prod_{i=1}^{t}\rho_{i}\big)\big)
\label{eq:nadam4}
\end{equation}

\begin{equation}
  \hat{u} \leftarrow v u_{t} / \big(1-v_t\big)
\label{eq:nadam5}
\end{equation}

\begin{equation}
  \theta_{t} \leftarrow \theta_{t-1} - \frac{\epsilon_{t}}{\sqrt{\hat{u}_{t}+\delta}} \hat{m}_{t}
\label{eq:nadam6}
\end{equation}

\subsection{AMSGrad}\label{ch:amsgrad}

Another an exponential moving average variant is AMSGrad \cite{reddi}. The purpose of developing AMSGrad is to guarantee the convergence while preserving the benefits of Adam and RMSProp. In AMSGrad algorithm, the first and second moment variables are updated as in Equations \ref{eq:amsgrad2} and \ref{eq:amsgrad3}. The key difference between AMSGrad and Adam is shown in Equation \ref{eq:amsgrad4}. Here, AMSGrad maintains the maximum of all {\it $u_{t}$} until the present time step and uses this maximum value for normalizing the running average of the gradient instead of {\it $u_{t}$} in Adam. By doing this, AMSGrad results in a non-increasing step size. Finally, the parameters are updated as in Equation \ref{eq:amsgrad6}. Here, $\hat{U}_{t}$ indicates $diag\big(\hat{u}_{t}\big)$. 

\begin{equation}
  m_{t} \leftarrow \rho_{1}^t m_{t-1} + (1-\rho_{1}^t)g_{t}
\label{eq:amsgrad2}
\end{equation}

\begin{equation}
  u_{t} \leftarrow \rho_2 u_{t-1} + (1-\rho_2)g_{t}^2
\label{eq:amsgrad3}
\end{equation}

\begin{equation}
  \hat{u}_{t} \leftarrow max\big(\hat{u}_{t-1},u_t \big) 
\label{eq:amsgrad4}
\end{equation} 

\begin{equation}
  \theta_{t+1} \leftarrow \prod\nolimits_{F,\sqrt{\hat{U}_{t}}} \big(\theta_t - \epsilon_t m_t / \sqrt{\hat{u}_{t}}\big)
\label{eq:amsgrad6}
\end{equation}

On the other side, the difference of AMSGrad from Adam and AdaGrad, it neither increases nor decreases the learning rate and furthermore, decreases {\it $u_{t}$} which can potentially lead to non-decreasing learning rate even if gradient is large in the future iterations \cite{reddi}.

\section{Experiments}\label{ch:experiments}

\subsection{Datasets}

The performances of optimization algorithms are evaluated on four image datasets, namely, MNIST \cite{lecun}, CIFAR-10 \cite{krizhevsky}, Kaggle Flowers \cite{mamaev} and Labeled Faces in the Wild (LFW) \cite{huang2}. The well-known dataset MNIST includes 60000 training and 10000 test examples each of which is a $28\times 28$ gray scale handwritten digit image. CIFAR-10 is composed of 50000 training and 10000 test examples. They are $32\times 32$ color images belonging to 10 classes. Kaggle Flowers contains 4242 images of 5 different types of flowers. LFW includes 13233 face images belonging to 5749 people. In supervised learning experiments, LFW classes that have at least 30 images are chosen. LFW has 34 classes meet this requirement, thus 1777 face images belong to 34 people are used for classification. On the other side, all the 13233 images are used in unsupervised learning experiments.  Besides, LFW and Kaggle flowers include color images in different sizes. In this study, LFW and Kaggle Flowers are scaled to a size of $64\times 64$ and $96\times 96$, respectively. Also, they are randomly divided into two subsets as 0.75 for training and 0.25 for testing.

\subsection{Experimental setting}

In this study, both supervised and unsupervised learning tasks are handled. Firstly, three different CNN architectures are used for classification task. First one includes three convolutional and two fully-connected (FC) layers similar to well-known LeNet-5 \cite{lecun}. Second one has five convolutional and three FC layers that are organized in the architecture similar to AlexNet \cite{krizhevsky2}. Last one has seven convolutional layers that are stacked VGG-style \cite{simonyan}. These architectures are shown in Table 1. All of them have convolutional layers that include filters with $3\times 3$ kernels in addition to $2\times 2$ max-pooling layers. In order to avoid overfitting, weight decay of $1e-4$ is applied to convolutional layers. Additionally, dropout in ratio 0.50 is applied to the FC layers preceding the output layer. All layers use ReLU as activation function except the output layer. The loss function is categorical cross entropy.

Secondly, convolutional autoencoders (CAE) are preferred for the unsupervised learning task. In general, a CAE comprises of two parts called encoder and decoder. While the encoder generates a representation of input data, the decoder takes this representation as input and reconstructs it in the output. When deciding the architecture of the autoencoder, the size of the representation is very important. Because when the size of representation becomes smaller, the reconstruction of the images becomes harder. Therefore, two architectures are preferred for the unsupervised learning experiments as described in Table 2. The encoder and decoder have symmetric convolutional-deconvolutional layers. Here, the difference between the encoder and decoder is that upsampling layers are used in decoder instead of max-pooling layers. The first architecture, CAE-1, reduces the size of representation to one quarter while the second one, CAE-2, reduces it to one half. For example, the representation of $64 \times 64$ images of LFW is 1024 dimensional in CAE-1 and 2048 dimensional in CAE-2. For unsupervised learning experiments, the reconstruction loss is mean squared error and the activation function is tanh in the output layer. The rest of the layers use ReLU. 

\begin{table}[th]
 \begin{minipage}{0.5\linewidth}
  \tbl{CNN architectures used in supervised learning experiments.}
  {\begin{tabular}{@{}ccc@{}} \toprule
  \textbf{CNN-1} & \textbf{CNN-2} & \textbf{CNN-3} \\
  Conv-32      & Conv-32      & Conv-32   \\                             
  MaxPool      & MaxPool      & Conv-32   \\
               &              & MaxPool    \\
  Conv-64      & Conv-64      &           \\
  MaxPool      & MaxPool      & Conv-64   \\
               &              & Conv-64   \\
  Conv-128     & Conv-128     & MaxPool   \\
  MaxPool      & Conv-128     &           \\
               & Conv-128     & Conv-128  \\
  FC-128       & MaxPool      & Conv-128  \\
  FC-softmax   &              & Conv-128  \\
               & FC-128       & MaxPool    \\
               & FC-256       &           \\
               & FC-softmax   & FC-128     \\                        
               &              & FC-256     \\
               &              & FC-softmax  \\ \botrule
  \end{tabular}}
  \label{cnn_architectures}  
 \end{minipage}%
\begin{minipage}{0.55\linewidth}
  \tbl{CAE architectures used in unsupervised learning experiments.} 
  {\begin{tabular}{p{2cm}p{2cm}}  \toprule   %
  \textbf{CAE-1} & \textbf{CAE-2}    \\ 
  
  Conv-32      & Conv-32        \\     
  MaxPool      & MaxPool      \\    
                          &                 \\      
  Conv-4       & Conv-8           \\
  MaxPool      & MaxPool         \\
               &                 \\
  Conv-4       & Conv-8          \\
  UpSample     & UpSample        \\
               &                 \\
  Conv-32      & Conv-32          \\
  UpSample     & UpSample          \\
               &                   \\
  Conv-tanh    & Conv-tanh          \\
               &                 \\ 
               &                 \\                                    
               &               \\ \botrule 
   \end{tabular}}
  \label{cae_architectures}  
 \end{minipage} 
\end{table}

In all experiments, the learning rate $\epsilon$ is 0.01 for SGD and its momentum variants; 0.001 for the algorithms with adaptive learning rates. The same learning rate is used for all algorithms, except SGD and its momentum variants. Because they do not perform well if the learning rate is smaller. Therefore, the hyperparameters which the algorithms give best results on test data are preferred. The momentum $\alpha$ is 0.9 for SGD variants. The decay constant $\rho$ is 0.95 for AdaDelta. Also, the decay constants $\rho_1$ and $\rho_2$ are 0.9 and 0.999 for Adam, AdaMax, Nadam and AMSGrad. The minibatch size is 128 for all models and each of them is trained 50 epochs. In order to compare different algorithms, the same parameter initialization is used. The software is based on Theano \cite{bergstra} on a single GTX 1050 Ti GPU.

\subsection{Supervised learning experiments}

\begin{figure}[htbp]  
\centerline{\includegraphics[width=5.2cm]{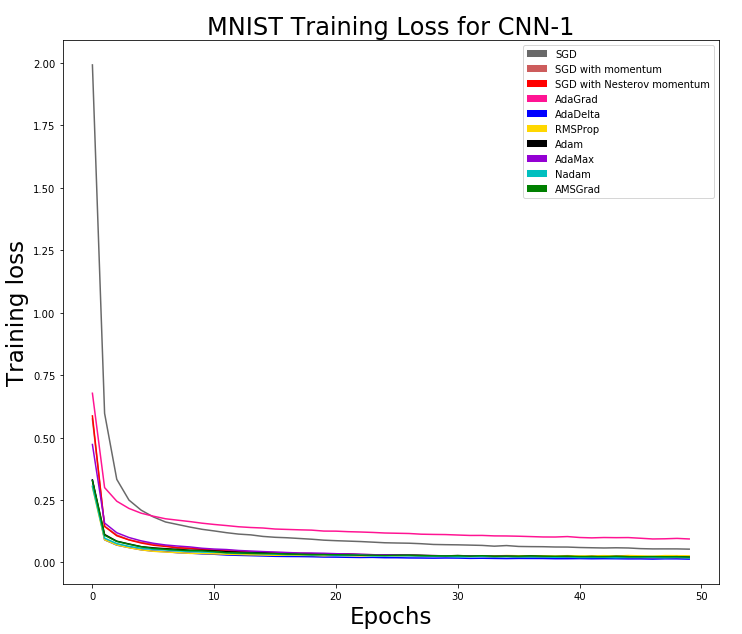} \includegraphics[width=5.2cm]{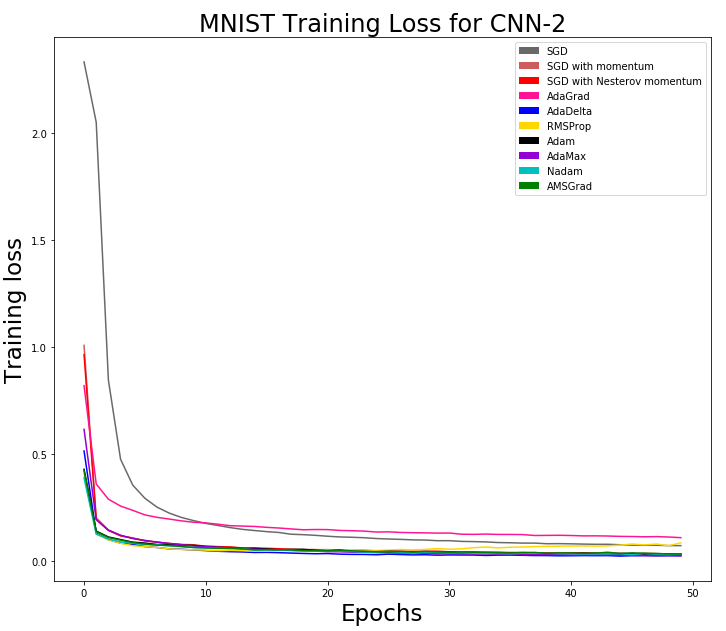} \includegraphics[width=5.2cm]{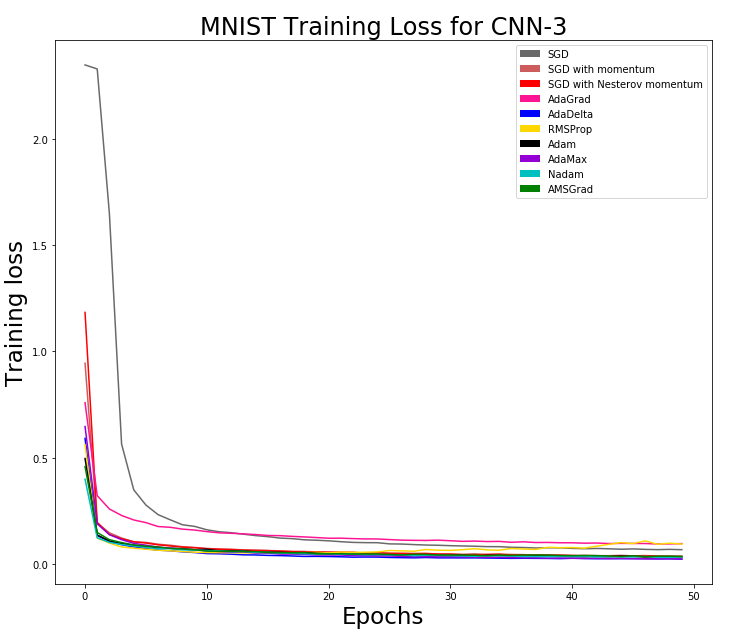} }
\vspace*{8pt}
\caption{The behaviour of algorithms on MNIST during training for three CNN architectures.}
\label{mnist_training_loss}
\end{figure}

In order to compare the optimization algorithms, all datasets are classified for each CNN architecture. The classification results for the basic dataset MNIST are given in Table 3. For all architectures, AdaDelta and AdaMax give the best accuracies on test data. On the other side, SGD with Nesterov momentum follows AdaDelta and AdaMax so closely for CNN-3. The worst performance belongs to AdaGrad which is the simplest adaptive learning algorithm. The behaviour of the algorithms during training is shown in Figure \ref{mnist_training_loss}. SGD begins training with the highest loss for all three architectures while Nadam begins with the lowest. Therefore, it can be said that Nadam makes a better start the training by using the same parameter initialization. Adam and its variants give similar results to each other for CNN-1. However, the difference between their performances is observed when the network becomes deeper. Besides, AdaDelta performs better than AdaGrad and RMSProp during training.   

\begin{table}[th]
\tbl{Comparison of the algorithms on MNIST for classification.}
{\begin{tabular}{ll cc c cc c l} \toprule
 
&  \multicolumn{2}{c}{\textbf{CNN-1}} & \multicolumn{2}{c}{\textbf{CNN-2}} & \multicolumn{2}{c}{\textbf{CNN-3}}\\
   
\textbf{Algorithm} &  \textbf{Test Loss} & \textbf{Test Acc.} & \textbf{Test Loss} & \textbf{Test Acc.} & \textbf{Test Loss} & \textbf{Test Acc.} \\ 
   
SGD & \hphantom{00}0.0468 & 99.00\hphantom{0} & \hphantom{0}0.0724 & 99.03 & \hphantom{0}0.0780 & 99.24   
\\
SGD - momentum & \hphantom{00}0.0395 & 99.26\hphantom{0} & \hphantom{0}0.0633 & 99.18 & \hphantom{0}0.0718 & 99.26 
\\
SGD - Nesterov  & \hphantom{00}0.0366 & 99.34\hphantom{0} & \hphantom{0}0.0511 & 99.35 & \hphantom{0}0.0589 & 99.41 
\\
AdaGrad & \hphantom{00}0.0600 & 98.57\hphantom{0} & \hphantom{0}0.0753 & 98.79 & \hphantom{0}0.0730 & 99.08 
\\
AdaDelta & \hphantom{00}0.0306 & \textbf{99.50}\hphantom{0} & \hphantom{0}0.0395 & 99.49 & \hphantom{0}0.0460 & 99.40 
\\
RMSProp & \hphantom{00}0.0505 & 99.26\hphantom{0} & \hphantom{0}0.1899 & 98.70 & \hphantom{0}0.1108 & 99.32 
\\
Adam & \hphantom{00}0.0425 & 99.35\hphantom{0} & \hphantom{0}0.0597 & 99.00 & \hphantom{0}0.0536 & 99.29 
\\
AdaMax & \hphantom{00}0.0337 & 99.37\hphantom{0} & \hphantom{0}0.0418 & \textbf{99.51} & \hphantom{0}0.0454 & \textbf{99.42} 
\\
Nadam & \hphantom{00}0.0364 & 99.32\hphantom{0} & \hphantom{0}0.0567 & 99.19 & \hphantom{0}0.0565 & 99.16 
\\
AMSGrad & \hphantom{00}0.0401 & 99.24\hphantom{0} & \hphantom{0}0.0561 & 99.28 & \hphantom{0}0.0497 & 99.36 \\ \botrule
\end{tabular}}
\label{mnist_supervised_results}
\end{table}

\begin{figure}[htbp]  
\centerline{\includegraphics[width=5.2cm]{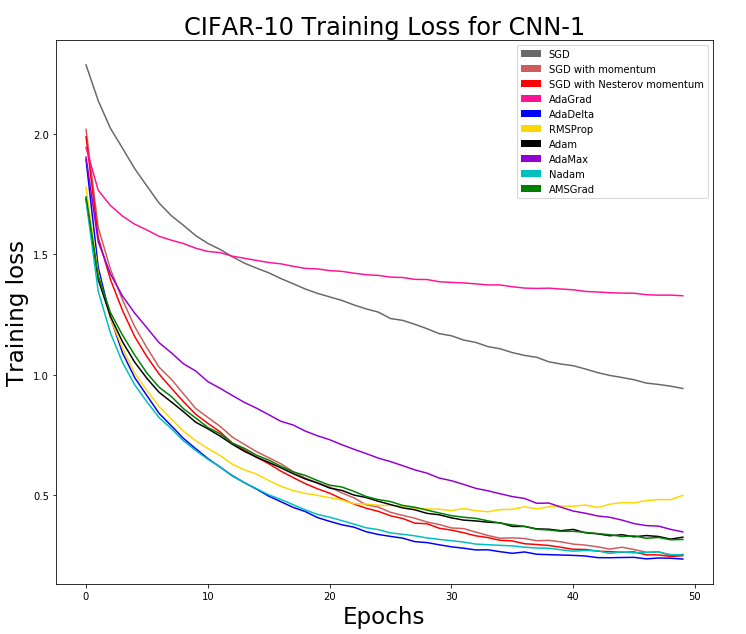} \includegraphics[width=5.2cm]{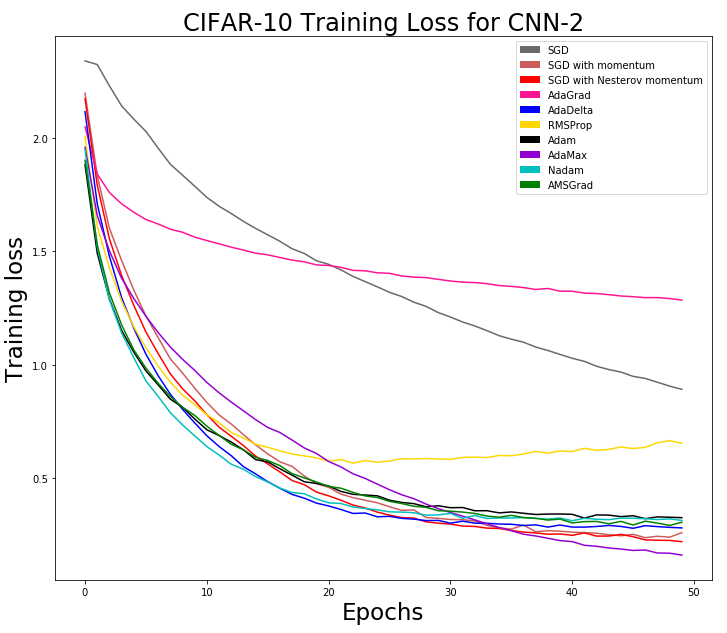} \includegraphics[width=5.2cm]{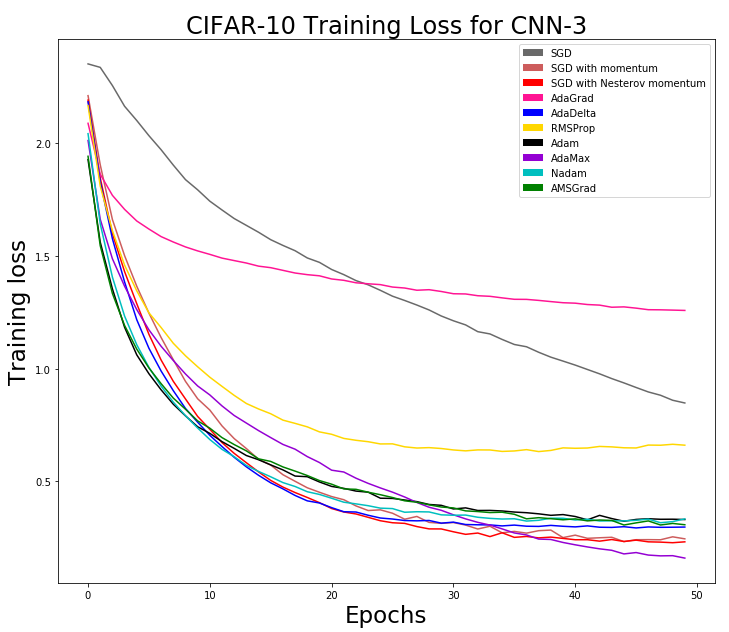} }
\vspace*{8pt}
\caption{The behaviour of algorithms on CIFAR-10 during training for three CNN architectures.}
\label{cifar10_training_loss}
\end{figure}

The classification results for CIFAR-10 are given in Table 4. As CIFAR-10 includes color images in higher resolutions than MNIST, the performances of algorithms begin to change. AdaMax gives the best accuracies on test data for all architectures. AMSGrad performs close to AdaMax for CNN-1 and CNN-2. On the other side, Adam and SGD with Nesterov momentum follow AdaMax when the neural network becomes deeper. Similar to MNIST results, AdaGrad can not perform well on CIFAR-10. The behaviour of the algorithms during training is shown in Figure \ref{cifar10_training_loss}. While SGD begins training with the highest loss, Adam begins with the lowest. Also, AdaMax seems to perform better for deeper architectures. 

\begin{table}[th]
\tbl{Comparison of the algorithms on CIFAR-10 for classification.}
{\begin{tabular}{ll cc c cc c l} \toprule
 
&  \multicolumn{2}{c}{\textbf{CNN-1}} & \multicolumn{2}{c}{\textbf{CNN-2}} & \multicolumn{2}{c}{\textbf{CNN-3}}\\
   
\textbf{Algorithm} &  \textbf{Test Loss} & \textbf{Test Acc.} & \textbf{Test Loss} & \textbf{Test Acc.} & \textbf{Test Loss} & \textbf{Test Acc.} \\ 
   
SGD & \hphantom{00}0.9428 & 67.98\hphantom{0} & \hphantom{0}0.8899 & 70.84 & \hphantom{0}1.0112 & 68.19         
\\
SGD - momentum & \hphantom{00}1.2479 & 74.96\hphantom{0} & \hphantom{0}1.3368 & 76.74 & \hphantom{0}1.1978 & 78.21 
\\
SGD - Nesterov & \hphantom{00}1.2235 & 76.01\hphantom{0} & \hphantom{0}1.2909 & 77.96 & \hphantom{0}1.1651 & 79.97 
\\
AdaGrad & \hphantom{00}1.2428 & 56.62\hphantom{0} & \hphantom{0}1.1769 & 59.61 & \hphantom{0}1.1531 & 60.55 
\\
AdaDelta & \hphantom{00}1.6619 & 74.08\hphantom{0} & \hphantom{0}1.7222 & 76.23 & \hphantom{0}1.4789 & 78.55 
\\
RMSProp & \hphantom{00}1.2670 & 75.01\hphantom{0} & \hphantom{0}1.1354 & 76.14 & \hphantom{0}0.9821 & 74.81 
\\
Adam & \hphantom{00}1.2279 & 76.05\hphantom{0} & \hphantom{0}1.1988 & 76.94 & \hphantom{0}0.9393 & 79.03 
\\
AdaMax & \hphantom{00}0.8054 & \textbf{76.89}\hphantom{0} & \hphantom{0}1.1240 & \textbf{79.17} & \hphantom{0}1.0437 & \textbf{81.41} 
\\
Nadam & \hphantom{00}1.2735 & 76.11\hphantom{0} & \hphantom{0}1.3102 & 76.60 & \hphantom{0}1.0672 & 78.98 
\\
AMSGrad & \hphantom{00}1.1940 & 76.45\hphantom{0} & \hphantom{0}1.2000 & 77.11 & \hphantom{0}1.0135 & 78.27 \\ \botrule
\end{tabular}}
\label{cifar10_supervised_results}
\end{table}

By now, the algorithms perform well except SGD and AdaGrad. The change of depth in neural network architectures does not significantly affect performance rankings of the algorithms. Therefore, the algorithms are compared on a more complicated dataset than MNIST and CIFAR-10. LFW consists of face images, each of which is a $64 \times 64$ color image belonging to one of 34 people. It includes more classes than MNIST and CIFAR-10 as well as higher resolutions. Also, it has small number of samples per each class. This classification task is more difficult for the optimization algorithms. The classification results for LFW are given in Table 5. The most obvious result is the decrease of the SGD performance. It can not learn nearly anything during training. Accordingly, its accuracy on test data is the lowest. Similarly, AdaGrad also can not perform well. The best accuracies on test data are obtained by using Adam and RMSProp for CNN-1, AdaDelta for CNN-2 and Adam for CNN-3. In the previous experiments, the performances of SGD momentum variants compete with the adaptive algorithms while training the neural networks on MNIST and CIFAR-10. However, while training the neural networks on LFW, the difference between them appears and the superiority of the adaptive learning algorithms become clear. Especially AdaDelta, Adam and its variants perform much better. 

\begin{table}[th]
\tbl{Comparison of the algorithms on LFW for classification.}
{\begin{tabular}{ll cc c cc c l} \toprule
 
&  \multicolumn{2}{c}{\textbf{CNN-1}} & \multicolumn{2}{c}{\textbf{CNN-2}} & \multicolumn{2}{c}{\textbf{CNN-3}}\\
   
\textbf{Algorithm} &  \textbf{Test Loss} & \textbf{Test Acc.} & \textbf{Test Loss} & \textbf{Test Acc.} & \textbf{Test Loss} & \textbf{Test Acc.} \\ 
   
SGD & \hphantom{00}2.8016 & 22.76\hphantom{0} & \hphantom{0}3.1141 & 19.22 & \hphantom{0}3.1152 & 19.22         
\\
SGD - momentum & \hphantom{00}1.0460 & 74.70\hphantom{0} & \hphantom{0}1.1003 & 71.66 & \hphantom{0}1.0803 & 73.18 
\\
SGD - Nesterov & \hphantom{00}0.9470 & 78.24\hphantom{0} & \hphantom{0}0.8571 & 79.25 & \hphantom{0}0.9930 & 75.37 
\\
AdaGrad & \hphantom{00}1.6427 & 58.17\hphantom{0} & \hphantom{0}1.8106 & 52.44 & \hphantom{0}1.7973 & 52.95 
\\
AdaDelta & \hphantom{00}0.8116 & 82.29\hphantom{0} & \hphantom{0}0.6383 &  \textbf{84.31} & \hphantom{0}0.8417 & 80.10 
\\
RMSProp & \hphantom{00}0.7532 &  \textbf{84.48}\hphantom{0} & \hphantom{0}0.7469 & 81.78 & \hphantom{0}0.9060 & 75.54 
\\
Adam & \hphantom{00}0.6887 &  \textbf{84.48}\hphantom{0} & \hphantom{0}0.8219 & 79.25 & \hphantom{0}0.7223 &  \textbf{82.96} 
\\
AdaMax & \hphantom{00}0.8734 & 77.74\hphantom{0} & \hphantom{0}0.9822 & 74.70 & \hphantom{0}0.8617 & 78.58 
\\
Nadam & \hphantom{00}0.7637 & 83.81\hphantom{0} & \hphantom{0}0.7172 & 82.96 & \hphantom{0}0.8015 & 80.26 
\\
AMSGrad & \hphantom{00}0.6795 & 82.63\hphantom{0} & \hphantom{0}0.7648 & 82.12 & \hphantom{0}0.6726 & 81.78 \\ \botrule
\end{tabular}}
\label{lfw_supervised_results}
\end{table}

\begin{figure}[htbp]  
\centerline{\includegraphics[width=5.2cm]{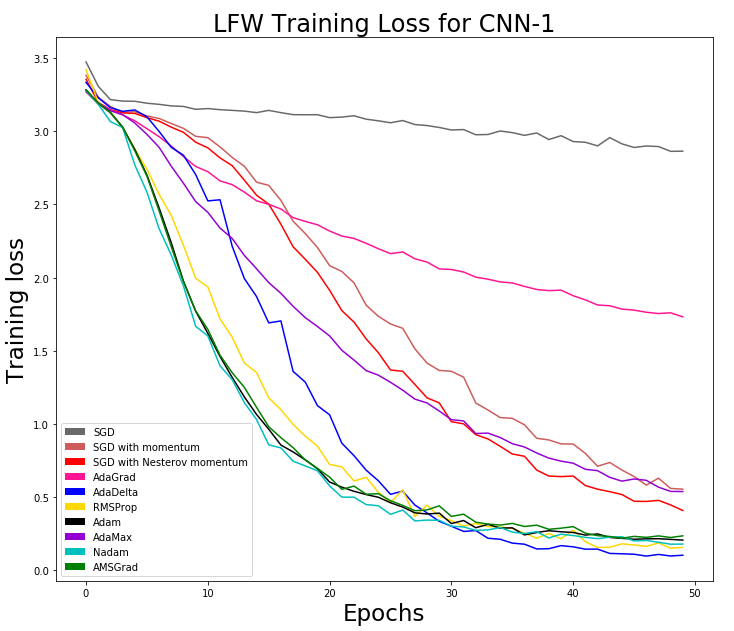} \includegraphics[width=5.2cm]{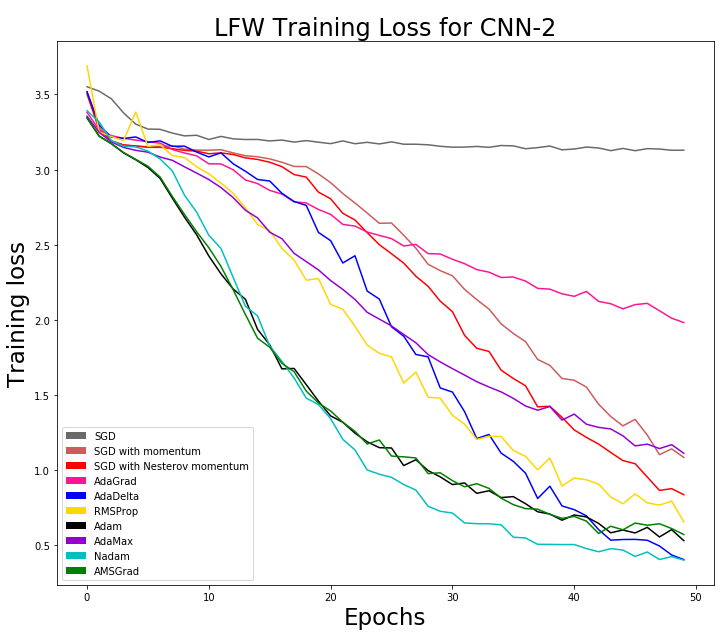} \includegraphics[width=5.2cm]{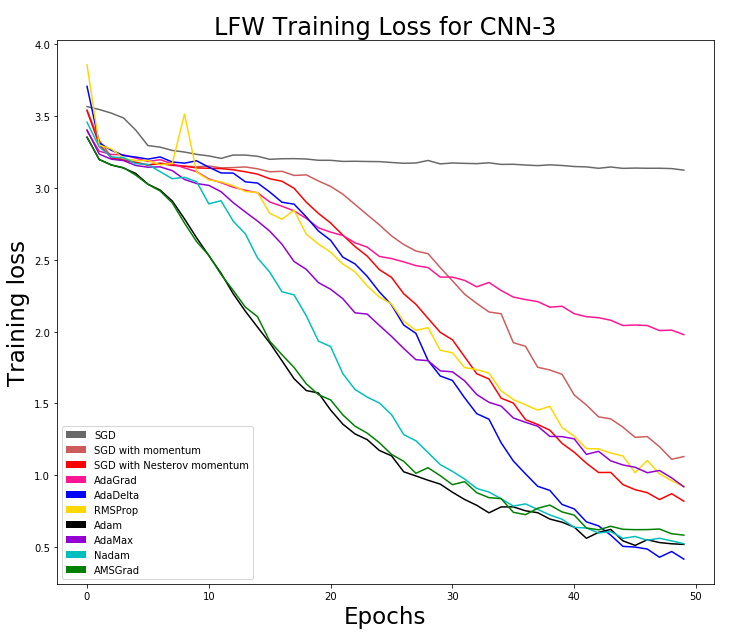} }
\vspace*{8pt}
\caption{The behaviour of algorithms on LFW during training for three CNN architectures.}
\label{lfw_training_loss}
\end{figure}
\newpage
The behaviour of algorithms during training is shown in Figure \ref{lfw_training_loss}. While SGD begins training with the highest loss for CNN-1, RMSProp begins with the highest for CNN-2 and CNN-3. On the other side, AdaDelta begins training with a high training loss but ends being one of lowest by giving a good accuracy on test data. As the network becomes deeper, RMSProp becomes worse on both training loss and test accuracy. Also, RMSProp has sometimes sudden peaks during training. This behaviour is more obvious for the deeper networks. While the training loss of Adam and AMSGrad are so close to each other, AdaMax falls behind its variants. In general, AdaDelta, Adam, Nadam and AMSGrad are significantly performs better during training. This result can be easily seen especially when the depth of neural network increases.

\begin{figure}[htbp]  
\centerline{\includegraphics[width=5.2cm]{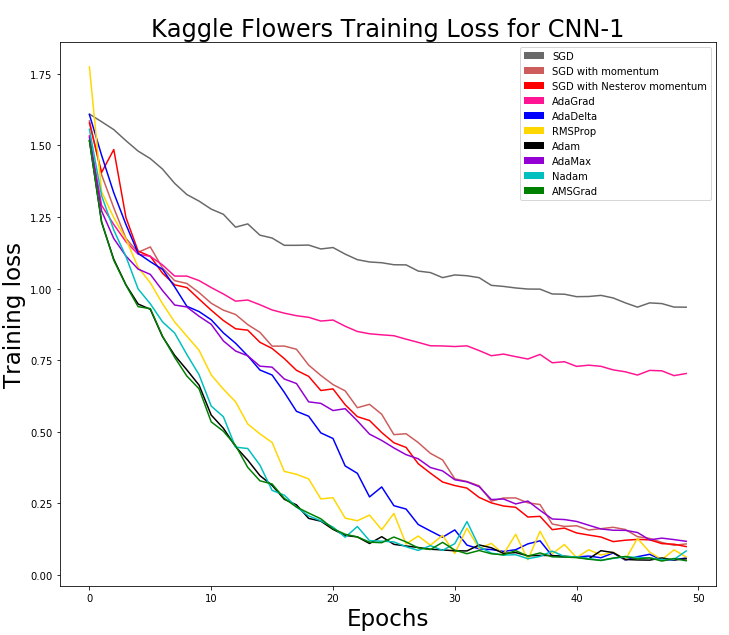} \includegraphics[width=5.2cm]{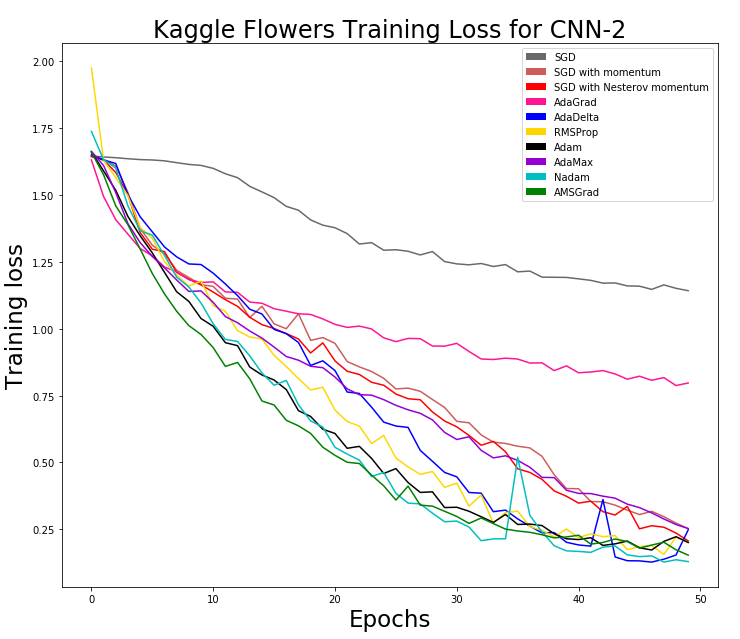} \includegraphics[width=5.2cm]{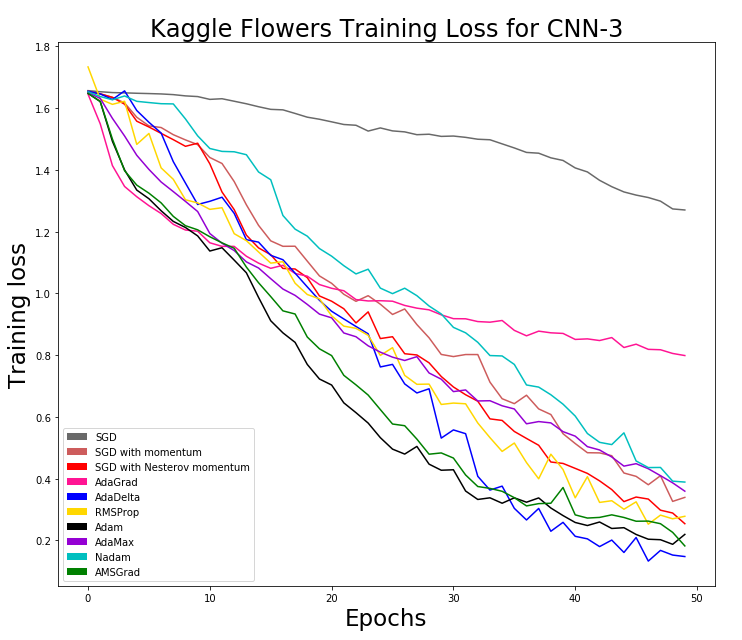} }
\vspace*{8pt}
\caption{The behaviour of algorithms on Kaggle Flowers during training for three CNN architectures.}
\label{kaggle_training_loss}
\end{figure}

Lastly, the performances of algorithms are compared on Kaggle Flowers which contains $96\times96$ images. The classification results for Kaggle Flowers are reported in Table 6 and the training process is shown in Figure \ref{kaggle_training_loss}. AdaMax and AMSGrad come into prominence according to accuracies on test data. On the contrary, SGD gives the worst results. RMSProp begins the training with the highest loss. However, the initial loss become closer for all algorithms when the neural network becomes deeper. Also, the performance of Nadam decreases when the depth increases. This behaviour of Nadam may arise because of the high resolution of images. AdaDelta, Nadam and RMSProp have sudden peaks during training. On the other side, AdaDelta is not so successful on test data even though it performs well during training. Interestingly, AdaGrad performs well especially for CNN-3. Its accuracy on test data is the closest result to the best which is performed by AdaMax. Therefore, AdaGrad seems to perform well for deeper networks as well as input images with high resolutions.   

\begin{table}[th]
\tbl{Comparison of the algorithms on Kaggle Flowers for classification.}
{\begin{tabular}{ll cc c cc c l} \toprule
 
&  \multicolumn{2}{c}{\textbf{CNN-1}} & \multicolumn{2}{c}{\textbf{CNN-2}} & \multicolumn{2}{c}{\textbf{CNN-3}}\\
   
\textbf{Algorithm} &  \textbf{Test Loss} & \textbf{Test Acc.} & \textbf{Test Loss} & \textbf{Test Acc.} & \textbf{Test Loss} & \textbf{Test Acc.} \\ 
   
SGD & \hphantom{00}1.0115 & 59.94\hphantom{0} & \hphantom{0}1.0764 & 59.66 & \hphantom{0}1.3027 & 44.40         
\\
SGD - momentum & \hphantom{00}1.5339 & 67.99\hphantom{0} & \hphantom{0}1.3234 & 69.19 & \hphantom{0}1.5486 & 63.64
\\
SGD - Nesterov & \hphantom{00}1.5711 & 68.54\hphantom{0} & \hphantom{0}1.4873 & 68.54 & \hphantom{0}1.7025 & 64.56 
\\
AdaGrad & \hphantom{00}0.8664 & 67.25\hphantom{0} & \hphantom{0}0.8639 & 66.60 & \hphantom{0}0.9359 & 66.88 
\\
AdaDelta & \hphantom{00}1.7850 & 68.64\hphantom{0} & \hphantom{0}1.8371 & 65.58 & \hphantom{0}2.1974 & 61.79
\\
RMSProp & \hphantom{00}1.9709 & 67.99\hphantom{0} & \hphantom{0}1.9051 & 66.88 & \hphantom{0}1.9653 & 63.55 
\\
Adam & \hphantom{00}1.9147 & 69.93\hphantom{0} & \hphantom{0}1.4274 & 68.82 & \hphantom{0}1.8564 & 62.16
\\
AdaMax & \hphantom{00}1.2073 & 69.93\hphantom{0} & \hphantom{0}1.1429 &  \textbf{71.32} & \hphantom{0}1.0537 &  \textbf{70.30} 
\\
Nadam & \hphantom{00}1.7687 & 67.06\hphantom{0} & \hphantom{0}1.9668 & 67.06 & \hphantom{0}1.4252 & 64.56 
\\
AMSGrad & \hphantom{00}1.8220 &  \textbf{70.67}\hphantom{0} & \hphantom{0}1.6430 &  \textbf{71.23} & \hphantom{0}1.9611 & 61.88 \\ \botrule
\end{tabular}}
\label{kaggle_flowers_supervised_results}
\end{table}

In supervised learning experiments, the positive effect of momentum for SGD is conspicuous especially for CIFAR-10, LFW and Kaggle Flowers datasets. Also, Nesterov momentum is slightly better despite of the momentum and Nesterov momentum results are close at the end of the training. In general, Adam, AdaMax and AMSGrad performs well on test data in addition to AdaDelta. Also, the most important advantage of AdaDelta is that it does not require to select the learning rate manually.  

\begin{figure}[htbp]  
\centerline{\includegraphics[width=5cm]{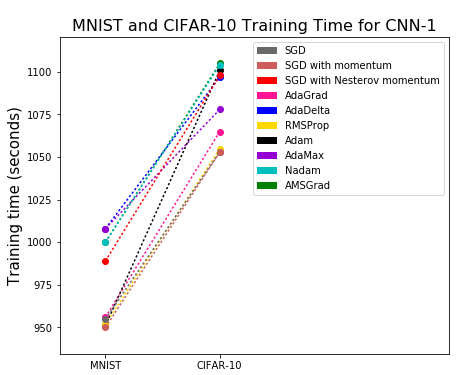} \includegraphics[width=5cm]{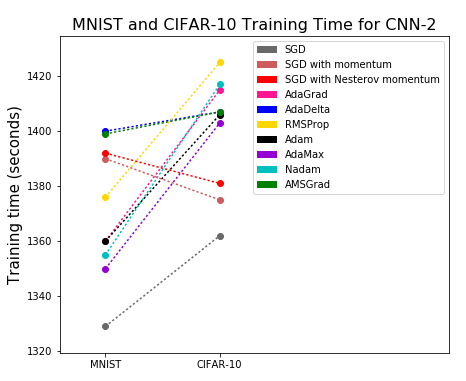} } \centerline{\includegraphics[width=5cm]{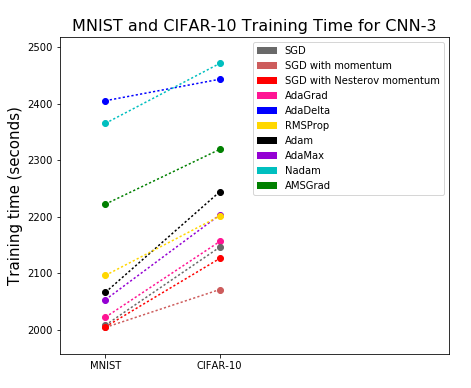}  \includegraphics[width=5cm]{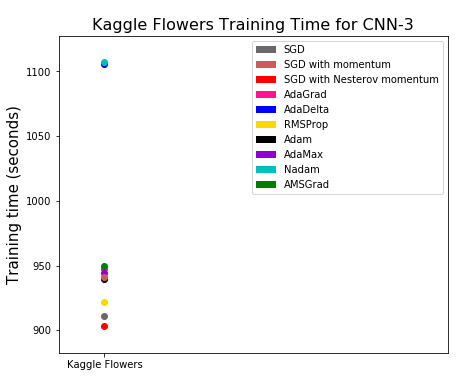} }
\vspace*{5pt}
\caption{{\it(Top)} MNIST and CIFAR-10 training time for CNN-1 and CNN-2. {\it(Bottom left)} MNIST training time for CNN-3. {\it(Bottom right)} Kaggle Flowers training time for CNN-3.}
\label{supervised_training_time}
\end{figure}

Additionally, the algorithms are compared based on the training time they require. As LFW includes a small number of training examples in supervised learning experiments, there is not significant difference between the training time of algorithms for each CNN architecture. On the other hand, MNIST and CIFAR-10 have more training examples than the other two datasets. Therefore, the training time vary across the algorithms as shown in Figure \ref{supervised_training_time}. For CNN-1, RMSProp is the fastest among the adaptive algorithms for both datasets. While AdaDelta and AdaMax are slow for MNIST, AMSGrad and Nadam require more time for CIFAR-10. When the dataset becomes more complex, Adam needs more time. AdaGrad performs fast but it can not generalize well on test data. Also, SGD and its momentum variants seem to be fast but their learning rate is bigger in addition to the smaller number of computational steps they have. Nevertheless, they can not generalize well similar to AdaGrad.

RMSProp, Nadam and AdaMax require more time on CIFAR-10 for CNN-2. However, the complexity of dataset does not matter for Adam and AMSGrad. When the neural network becomes deeper, in CNN-3, Nadam and AdaDelta take more time in comparison with the others for both datasets. In order to train CNN-1 and CNN-2, the difference between the slowest and fastest algorithms is less than 1 minutes. On the other hand, this required time increases to nearly 5 minutes for CNN-3. AdaMax takes minimum time among the Adam variants for the deepest CNN. Additionally, the training time differs when CNN-3 is trained on Kaggle Flowers. As seen in Figure \ref{supervised_training_time}, Nadam and AdaDelta take more time. RMSProp requires less time on Kaggle Flowers in comparison with the other adaptive algorithms.

\subsection{Unsupervised learning experiments}

When CAE is used, the task is to reconstruct input images in the output layer. When denoising CAE is used, noisy images are taken as input and the task is to reconstruct clean images in the output layer. Therefore, a gaussian noise matrix is applied to input images before using denoising CAE. In these experiments, the noise factor is 0.25. Both vanilla and denoising autoencoder results on MNIST are given in Table 7. While Adam gives the lowest reconstruction loss for CAE-1 architecture, AdaMax and Nadam give the lowest for CAE-2. But the performances of Adam and its variants are so close to each other. On the other side, Nadam gives the best results for denosing CAE. RMSProp performs well on denoising task in addition to Adam variants. The performance of AdaMax slightly decreases for denoising task. Besides, AdaGrad falls behind the other adaptive algorithms in both tasks. The training loss for MNIST is shown in Figure 6. The behavior of Adam and AMSGrad beginning the training with the lowest loss is in sight. 

The results on CIFAR-10, LFW and Kaggle Flowers are similar. Adam and its variants give the best results. Different from MNIST results, RMSProp begins training with the lowest reconstruction loss. However, Adam and its variants give better results at the end of the training. The results on Kaggle Flowers are interesting. For CAE-1, the performances of Adam and AMSGrad get worse in denosing task.  Accordingly, it can be said that when the size of representation is small, Adam and AMSGrad can not perform well on high resolution images. Also, RMSProp performs well together with Adam and its variants on Kaggle Flowers for both unsupervised tasks. Both vanilla and denoising autoencoder results are given in Table 8 for CIFAR-10, Table 9 for LFW and Table 10 for Kaggle Flowers, respectively. Also, the behaviour of the optimization algorithms during training for the unsupervised learning tasks on CIFAR-10 is shown in Figure 7, LFW in Figure 8 and Kaggle Flowers in Figure 9.

\begin{figure}[b]  
\centerline{\includegraphics[width=5cm]{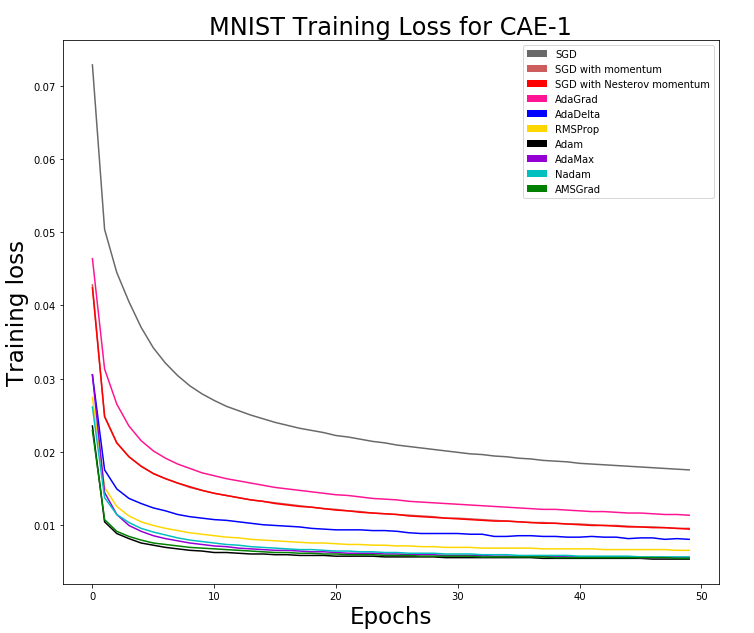} \includegraphics[width=5cm]{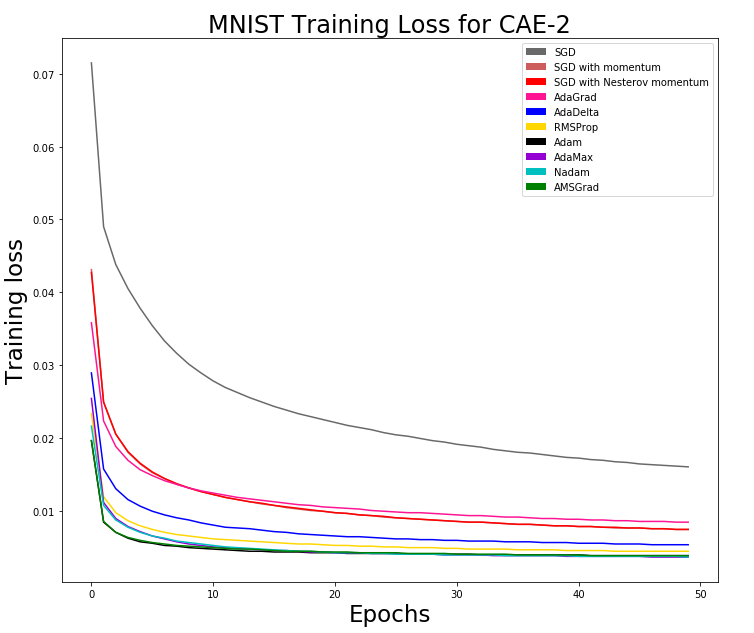} }
\centering
\includegraphics[width=5cm]{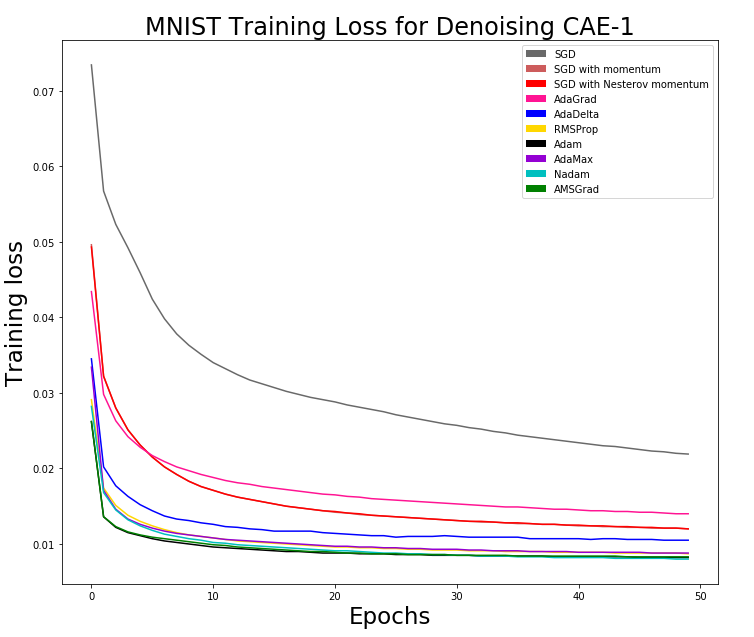} \includegraphics[width=5cm]{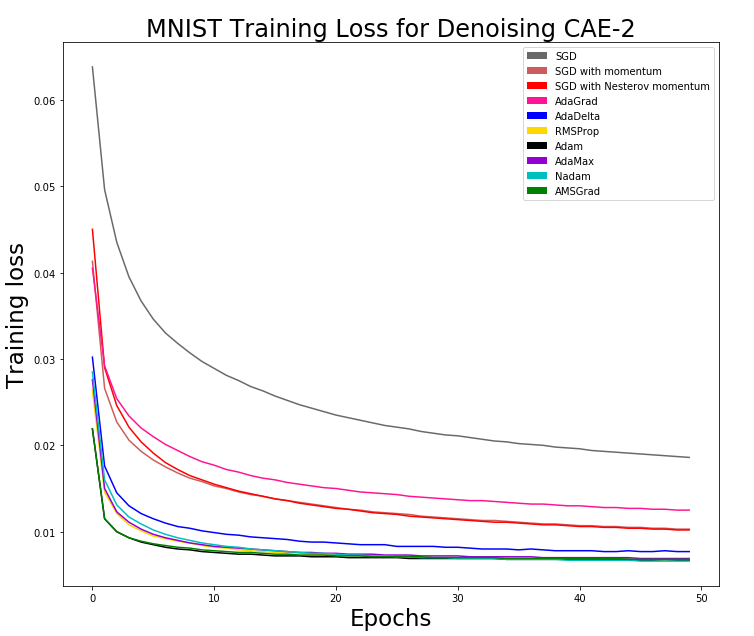}
\vspace*{5pt}
\caption{The behaviour of algorithms on MNIST during training for two autoencoder architectures. {\it(Top)} Training loss for CAE. {\it(Bottom)} Training loss for denoising CAE.}
\end{figure}

\begin{table}[ht]
\tbl{Comparison of the algorithms on MNIST for reconstruction.}
{\begin{tabular}{ll cc c cc c l} \toprule
 
&  \multicolumn{2}{c}{\textbf{CAE}} & \multicolumn{2}{c}{\textbf{Denoising CAE}} \\
   
\textbf{Algorithm} &  \textbf{CAE-1 Loss} & \textbf{CAE-2 Loss} & \textbf{CAE-1 Loss} & \textbf{CAE-2 Loss} \\ 
   
SGD & \hphantom{000}0.0175 & 0.0160 & 0.0219 & 0.0186    
\\
SGD - momentum & \hphantom{000}0.0095 & 0.0074 & 0.0120 & 0.0103 
\\
SGD - Nesterov  & \hphantom{000}0.0094 & 0.0074 & 0.0120 & 0.0102 
\\
AdaGrad & \hphantom{000}0.0113 & 0.0084 & 0.0140 & 0.0125 
\\
AdaDelta & \hphantom{000}0.0080 & 0.0053 & 0.0105 & 0.0077 
\\
RMSProp & \hphantom{000}0.0065 & 0.0044 & 0.0087 & 0.0067 
\\
Adam & \hphantom{000}\textbf{0.0053} & 0.0038 & 0.0082 & 0.0067  
\\
AdaMax & \hphantom{000}0.0055 & \textbf{0.0036} & 0.0088 & 0.0069  
\\
Nadam & \hphantom{000}0.0056 & \textbf{0.0036} & \textbf{0.0080} & \textbf{0.0066}  
\\
AMSGrad & \hphantom{000}0.0054 & 0.0038 & 0.0083 & 0.0068 \\ \botrule
\end{tabular}}
\label{cae_results_mnist}
\end{table}

\begin{table}[ht]
\tbl{Comparison of the algorithms on CIFAR-10 for reconstruction.}
{\begin{tabular}{ll cc c cc c l} \toprule
 
&  \multicolumn{2}{c}{\textbf{CAE}} & \multicolumn{2}{c}{\textbf{Denoising CAE}} \\
   
\textbf{Algorithm} &  \textbf{CAE-1 Loss} & \textbf{CAE-2 Loss} & \textbf{CAE-1 Loss} & \textbf{CAE-2 Loss} \\ 
   
SGD & \hphantom{000}0.0145 & 0.0152 & 0.0189 & 0.0158    
\\
SGD - momentum & \hphantom{000}0.0093 & 0.0091 & 0.0119 & 0.0108 
\\
SGD - Nesterov  & \hphantom{000}0.0093 & 0.0089 & 0.0119 & 0.0109 
\\
AdaGrad & \hphantom{000}0.0115 & 0.0067 & 0.0167 & 0.0113 
\\
AdaDelta & \hphantom{000}0.0093 & 0.0088 & 0.0113 & 0.0094 
\\
RMSProp & \hphantom{000}0.0072 & 0.0059 & 0.0102 & 0.0083 
\\
Adam & \hphantom{000}\textbf{0.0055} & 0.0047 & 0.0096 & 0.0075  
\\
AdaMax & \hphantom{000}0.0061 & 0.0047 & \textbf{0.0087} & 0.0074  
\\
Nadam & \hphantom{000}0.0069 & \textbf{0.0046} & \textbf{0.0087} & \textbf{0.0073}  
\\
AMSGrad & \hphantom{000}0.0060 & 0.0049 & 0.0098 & 0.0075 \\ \botrule
\end{tabular}}
\label{cae_results_cifar10}
\end{table} 

\begin{figure}[h]  
\centerline{\includegraphics[width=5cm]{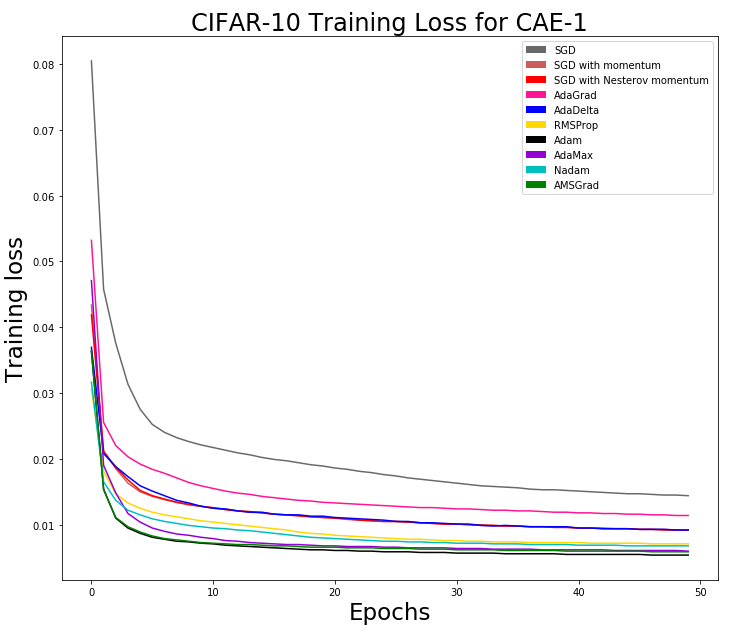} \includegraphics[width=5cm]{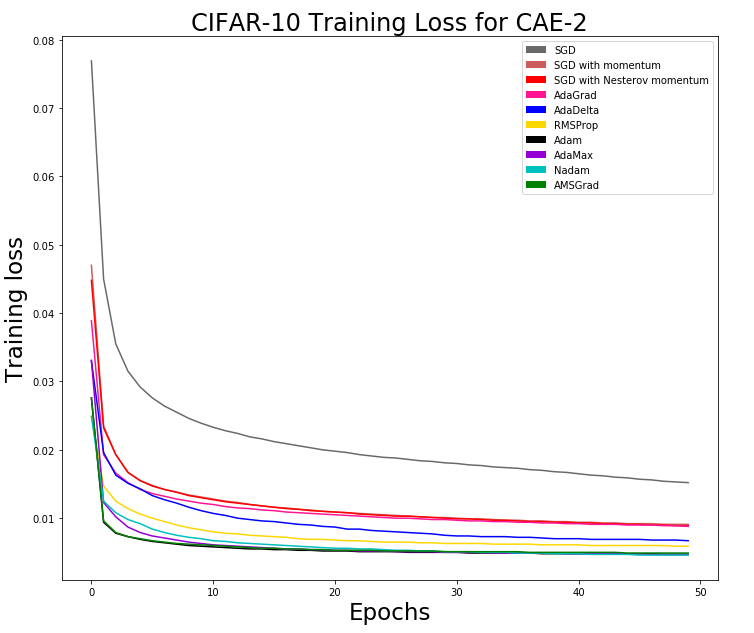} }
\centering
\includegraphics[width=5cm]{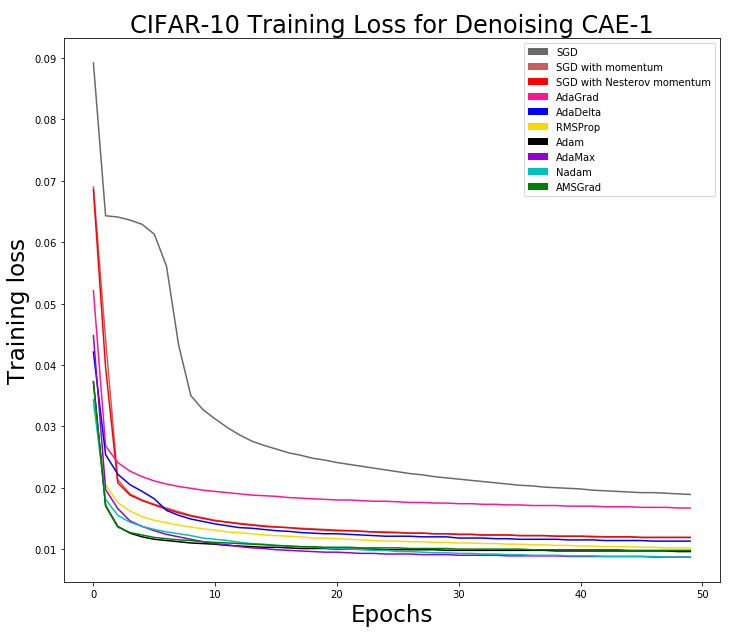} \includegraphics[width=5cm]{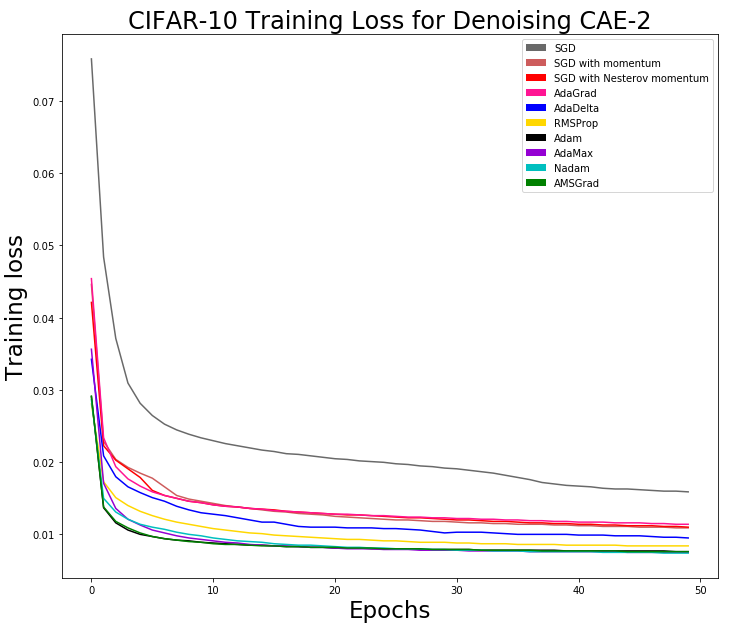}
\vspace*{8pt}
\caption{The behaviour of algorithms on CIFAR-10 during training for two autoencoder architectures. {\it(Top)} Training loss for CAE. {\it(Bottom)} Training loss for denoising CAE.}
\end{figure}

\begin{table}[t]
\tbl{Comparison of the algorithms on LFW for reconstruction.}
{\begin{tabular}{ll cc c cc c l} \toprule
 
&  \multicolumn{2}{c}{\textbf{CAE}} & \multicolumn{2}{c}{\textbf{Denoising CAE}} \\
   
\textbf{Algorithm} &  \textbf{CAE-1 Loss} & \textbf{CAE-2 Loss} & \textbf{CAE-1 Loss} & \textbf{CAE-2 Loss} \\ 
   
SGD & \hphantom{000}0.0122 & 0.0136 & 0.0125 & 0.0094    
\\
SGD - momentum & \hphantom{000}0.0070 & 0.0077 & 0.0073 & 0.0063 
\\
SGD - Nesterov  & \hphantom{000}0.0070 & 0.0077 & 0.0071 & 0.0063 
\\
AdaGrad & \hphantom{000}0.0071 & 0.0046 & 0.0090 & 0.0069 
\\
AdaDelta & \hphantom{000}0.0061 & 0.0055 & 0.0086 & 0.0066 
\\
RMSProp & \hphantom{000}0.0044 & 0.0033 & 0.0062 & 0.0053 
\\
Adam & \hphantom{000}\textbf{0.0030} & \textbf{0.0018} & \textbf{0.0044} & \textbf{0.0039}  
\\
AdaMax & \hphantom{000}0.0033 & 0.0019 & 0.0052 & 0.0043  
\\
Nadam & \hphantom{000}0.0036 & 0.0025 & 0.0050 & 0.0045  
\\
AMSGrad & \hphantom{000}\textbf{0.0030} & \textbf{0.0018} & 0.0049 & 0.0041 \\ \botrule
\end{tabular}}
\label{cae_results_lfw}
\end{table} 

\begin{figure}[t]  
\centerline{\includegraphics[width=4.7cm]{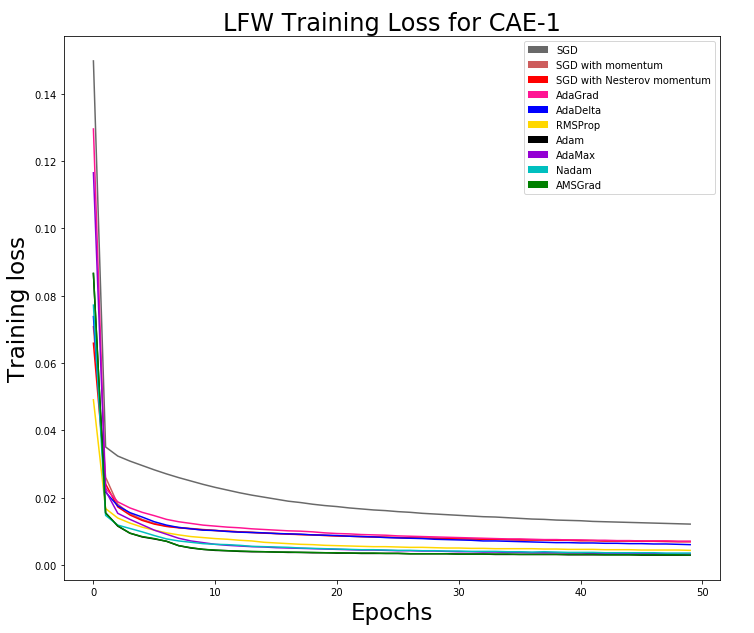} \includegraphics[width=4.7cm]{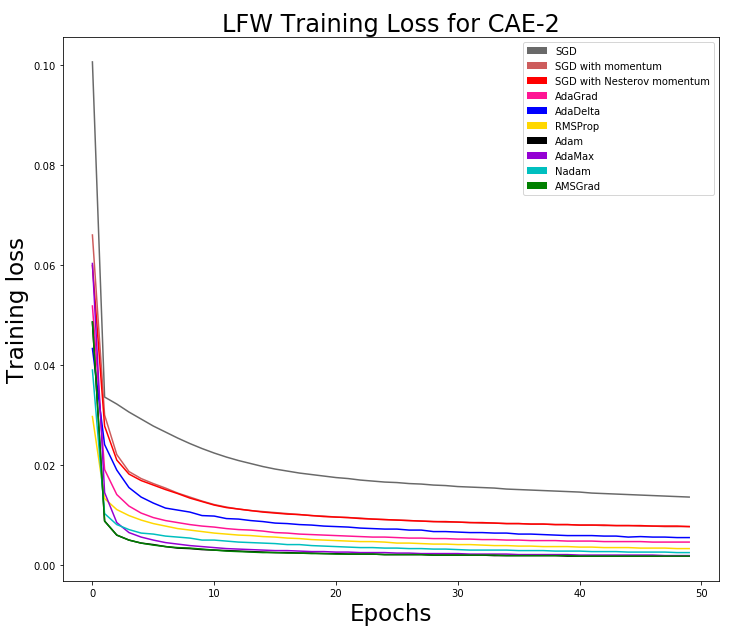} }
\centering
\includegraphics[width=4.7cm]{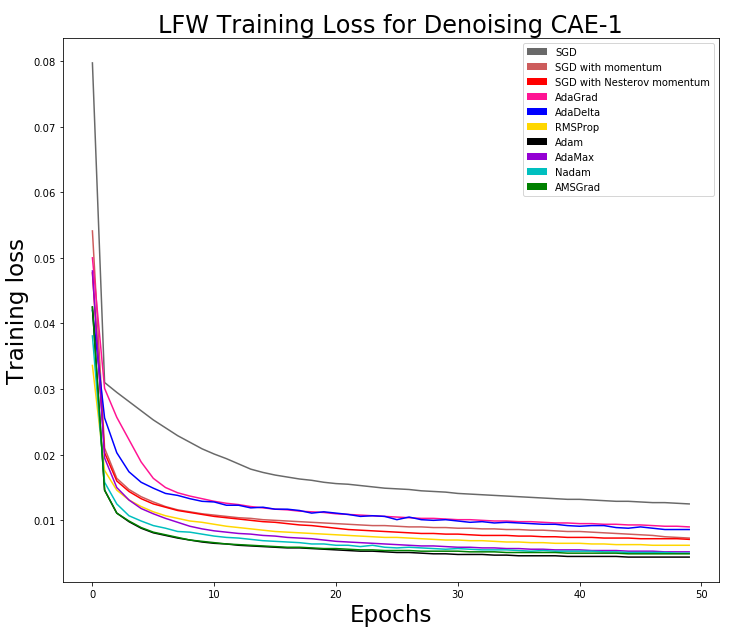} \includegraphics[width=4.7cm]{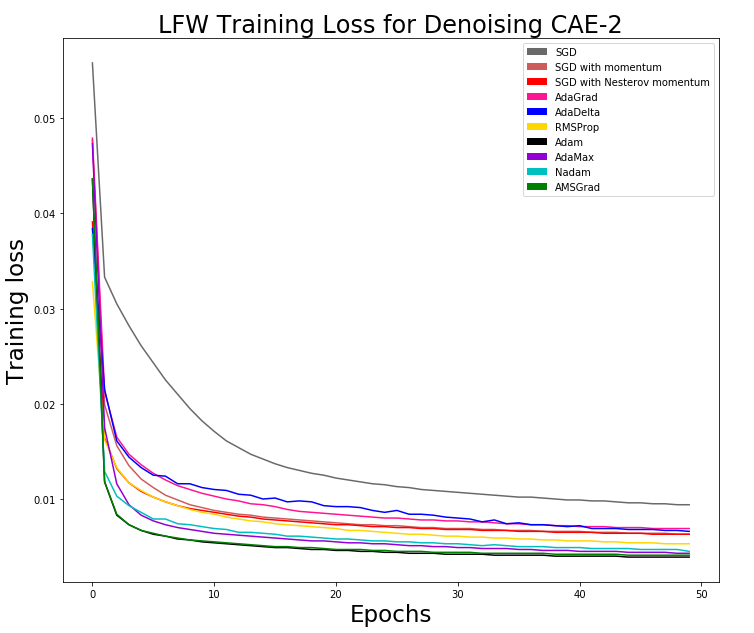}
\vspace*{8pt}
\caption{The behaviour of algorithms on LFW during training for two autoencoder architectures. {\it(Top)} Training loss for CAE. {\it(Bottom)} Training loss for denoising CAE.}
\end{figure}

\begin{table}[t] 
\tbl{Comparison of the algorithms on Kaggle Flowers for reconstruction.}
{\begin{tabular}{ll cc c cc c l} \toprule
 
&  \multicolumn{2}{c}{\textbf{CAE}} & \multicolumn{2}{c}{\textbf{Denoising CAE}} \\
   
\textbf{Algorithm} &  \textbf{CAE-1 Loss} & \textbf{CAE-2 Loss} & \textbf{CAE-1 Loss} & \textbf{CAE-2 Loss} \\ 
   
SGD & \hphantom{000}0.0359 & 0.0385 & 0.0367 & 0.0347   
\\
SGD - momentum & \hphantom{000}0.0217 & 0.0220 & 0.0320 & 0.0217 
\\
SGD - Nesterov  & \hphantom{000}0.0217 & 0.0214 & 0.0230 & 0.0219 
\\
AdaGrad & \hphantom{000}0.0298 & 0.0179 & 0.0254 & 0.0234 
\\
AdaDelta & \hphantom{000}0.0220 & 0.0180 & 0.0236 & 0.0203 
\\
RMSProp & \hphantom{000}0.0202 & 0.0159 & 0.0216 & 0.0179 
\\
Adam & \hphantom{000}\textbf{0.0144} & \textbf{0.0121} & 0.0269 & 0.0148  
\\
AdaMax & \hphantom{000}0.0205 & 0.0133 & 0.0209 & 0.0164  
\\
Nadam & \hphantom{000}0.0170 & 0.0138 & \textbf{0.0208} & 0.0166  
\\
AMSGrad & \hphantom{000}0.0145 & 0.0122 & 0.0270 & \textbf{0.0146} \\ \botrule
\end{tabular}}
\label{cae_results_kaggle_flowers}
\end{table} 

\begin{figure}[t]  
\centerline{\includegraphics[width=5cm]{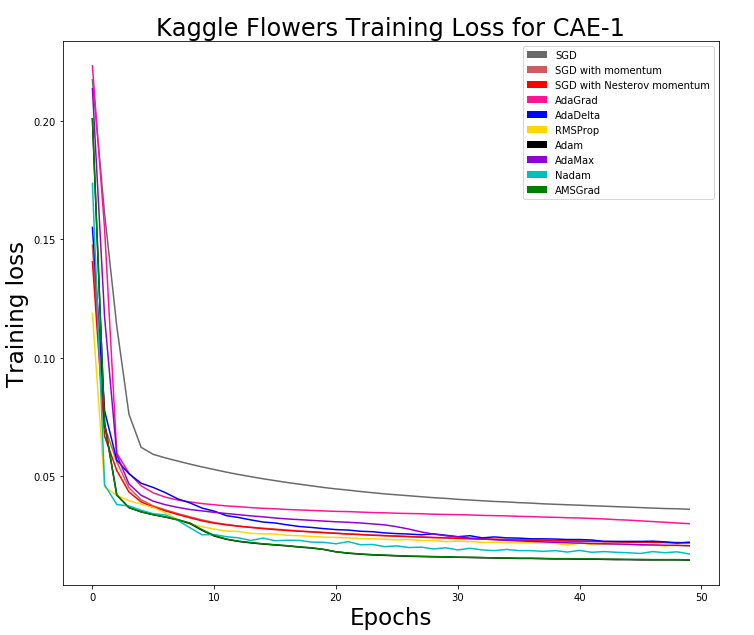} \includegraphics[width=5cm]{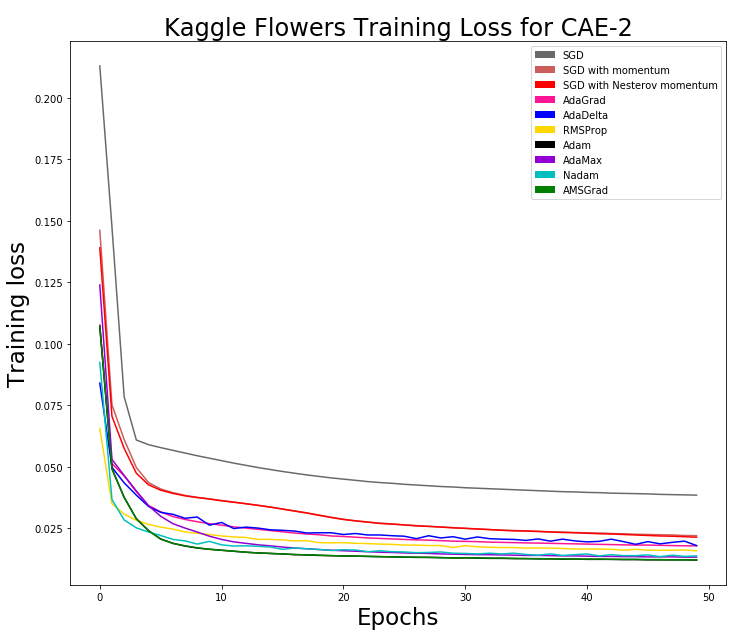} }
\centering
\includegraphics[width=5cm]{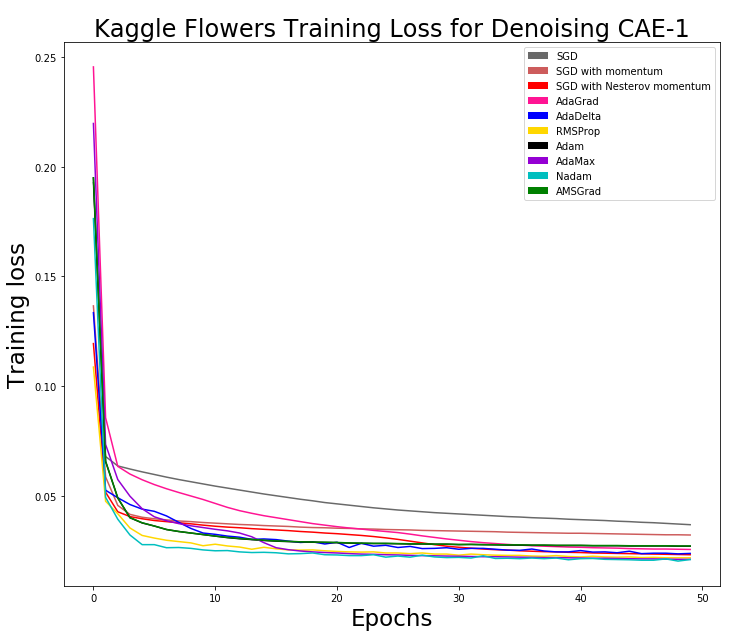} \includegraphics[width=5cm]{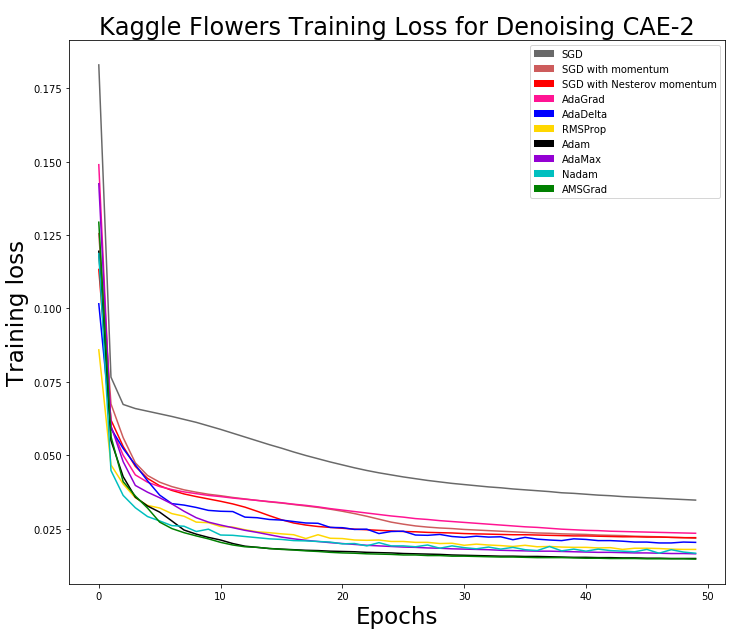}
\vspace*{8pt}
\caption{The behaviour of algorithms on Kaggle Flowers during training for two autoencoder architectures. {\it(Top)} Training loss for CAE. {\it(Bottom)} Training loss for denoising CAE.}
\end{figure}

SGD by far the worst for both unsupervised tasks on all datasets. In general, Adam and its variants seem to be much better than the other adaptive gradient methods in addition to SGD and its momentum variants. AdaGrad can not perform well among the adaptive algorithms.

\pagebreak 
In order to compare the results clearly in visual, the reconstructions of test images obtained by all algorithms for CAE-1 architecture are shown in Figure 10. As handwritten images are simpler than the other datasets, all algorithms perform well on MNIST. Even though the reconstructions of SGD are a little blurry, the digits can be seen clearly. An important point is based on the Kaggle Flowers results. SGD and AdaMax can not reconstruct colors properly. Similar results can be seen for CAE-2 as shown in Figure 11. Here, it seems that all adaptive algorithms can reconstruct images with their colors when the size of representation increases.

For denoising task, the reconstructions of all datasets for CAE-1 are given in Figure 12. The small size of representation makes this task more difficult. None of algorithms can reconstruct colors properly on Kaggle Flowers. Also, as it is hard to reconstruct faces and objects, the results are blurry. However, as the representation size increases, the colors given by especially Adam and its variants are much better as shown in Figure 13. Also, while SGD and AdaGrad can reconstruct images with their colors on CIFAR-10 for CAE-1, both algorithms fail in denosing CAE-1. 

\begin{figure}[th]  
\centerline{\includegraphics[width=12.4cm]{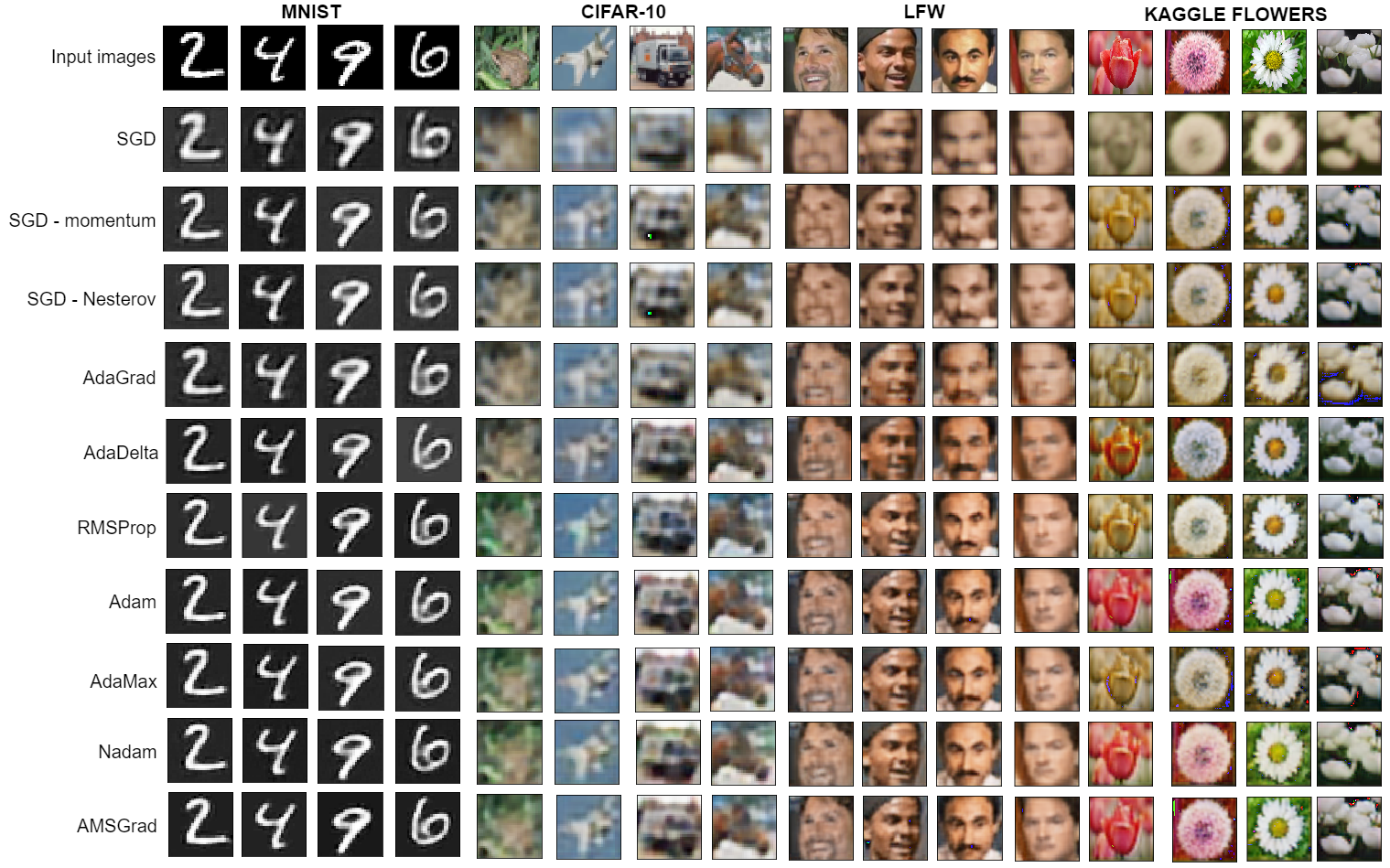}}
\vspace*{1pt}
\caption{Reconstructions of test images using CAE-1.}
\end{figure}

\begin{figure}[bh]  
\centerline{\includegraphics[width=12.4cm]{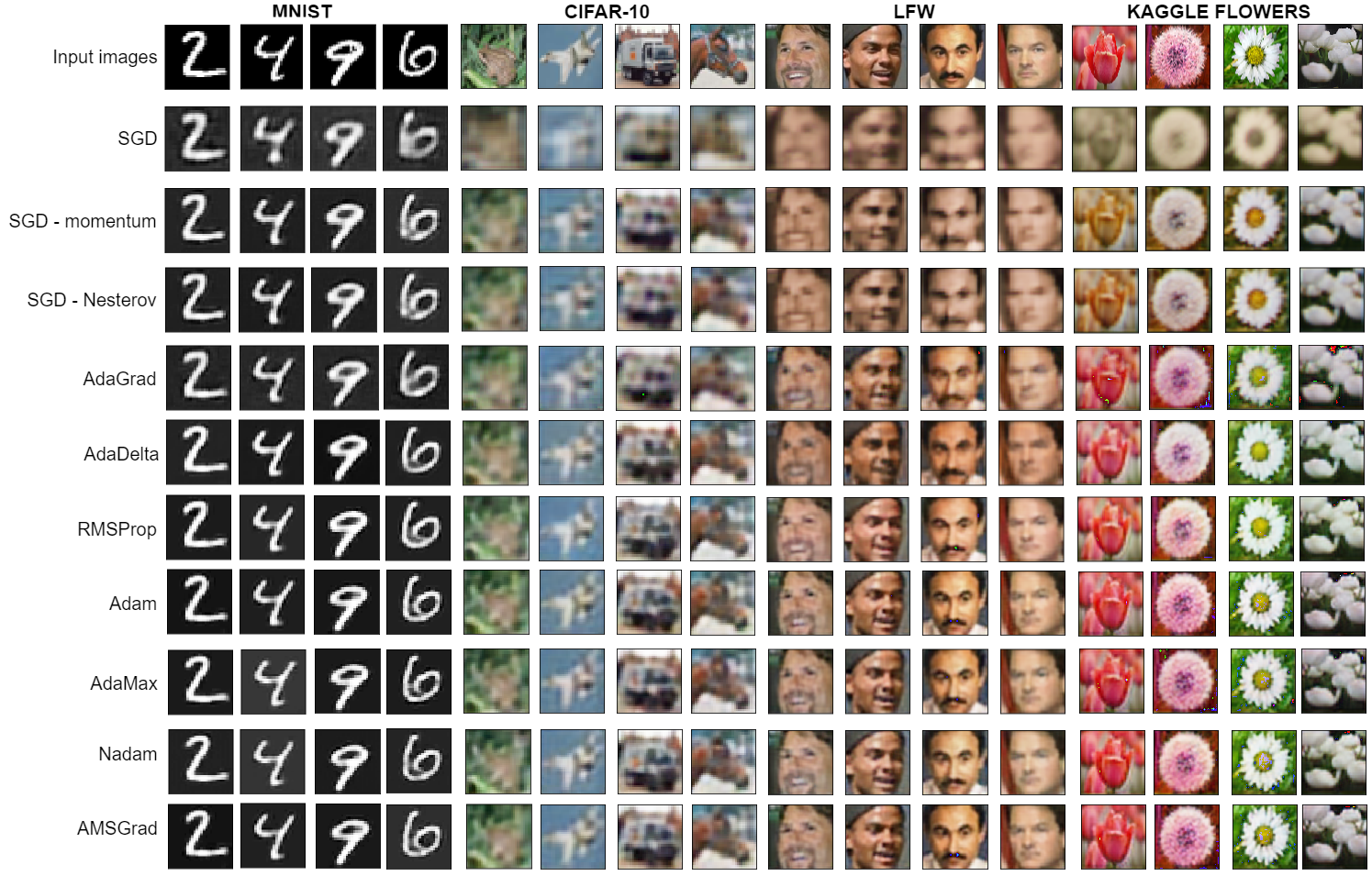}}
\vspace*{1pt}
\caption{Reconstructions of test images using CAE-2.}
\end{figure}

\begin{figure}[th]  
\centerline{\includegraphics[width=12.4cm]{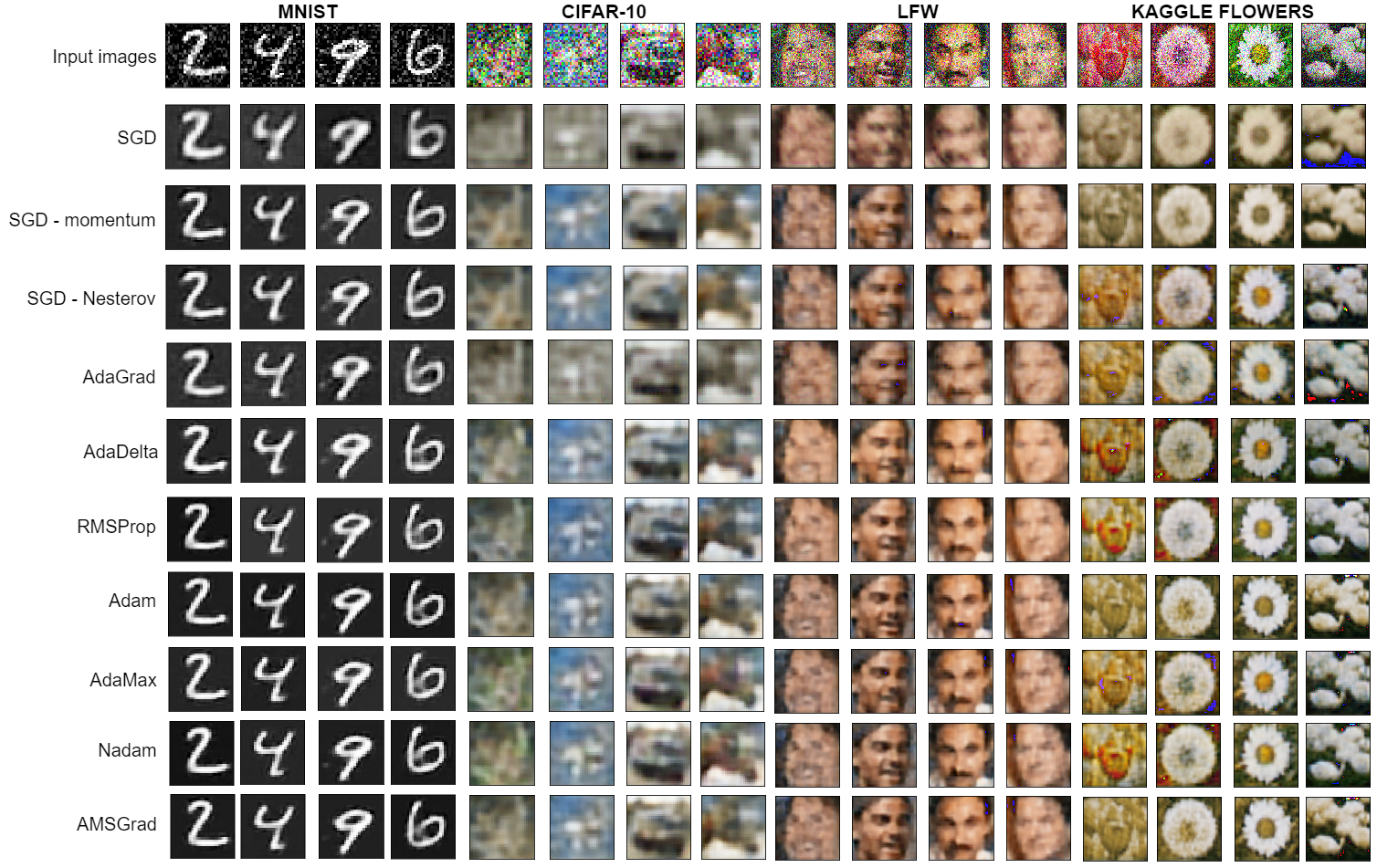}}
\vspace*{1pt}
\caption{Reconstructions of test images using denoising CAE-1.}
\end{figure}

\begin{figure}[bh]  
\centerline{\includegraphics[width=12.4cm]{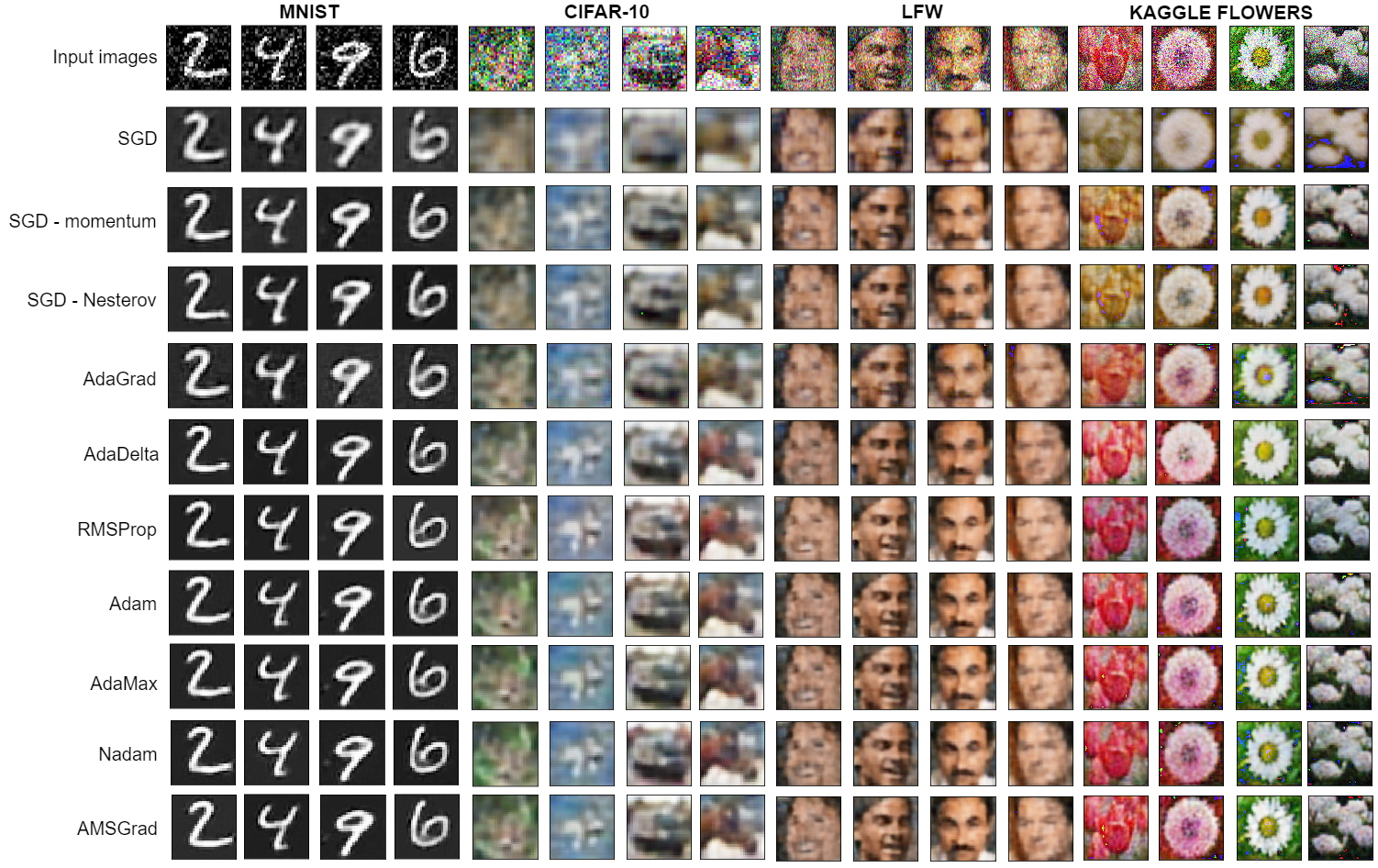}}
\vspace*{1pt}
\caption{Reconstructions of test images using denoising CAE-2.}
\end{figure}

Lastly, the algorithms are compared based on the training time they require. Similar to supervised learning experiments, training time does not vary significantly on LFW and Kaggle Flowers. However, training time on MNIST and CIFAR-10 for unsupervised learning experiments are shown in Figure 14. In general, Adam and Nadam need more time than other algorithms to reconstruct images. Similar to supervised results, when the dataset becomes more complex, they need more time. Also, AdaMax requires more time on CIFAR-10 as especially seen in CAE-1. All algorithms need more time for CAE-2 relative to CAE-1. Also, denoising task cause training time mostly decrease on CIFAR-10 and increase on MNIST. Even though SGD and its momentum variants train autoencoders fast by using the learning rate they perform best, they can not generalize well on test data.

\begin{figure}[h]  
\centerline{\includegraphics[width=5cm]{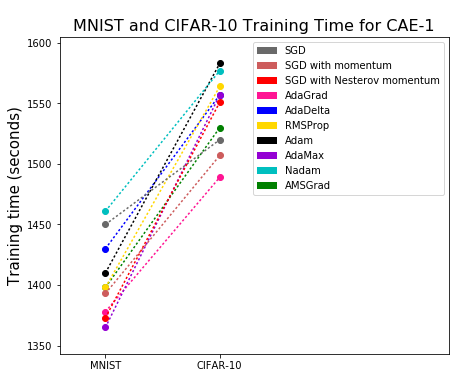} \includegraphics[width=5cm]{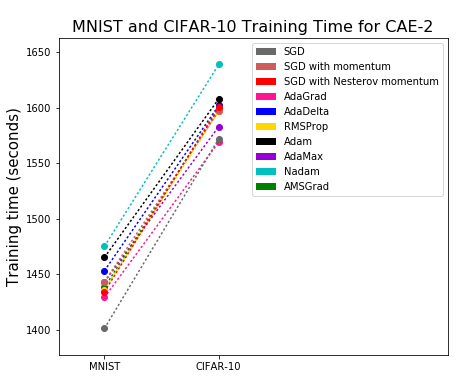} } \centerline{\includegraphics[width=5cm]{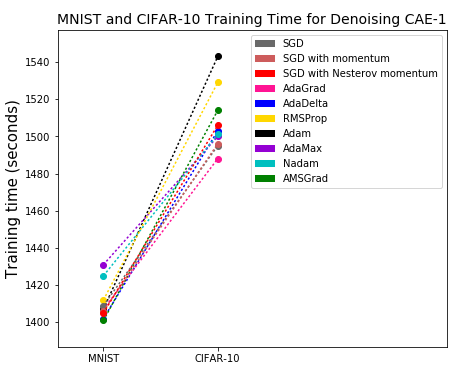} \includegraphics[width=5cm]{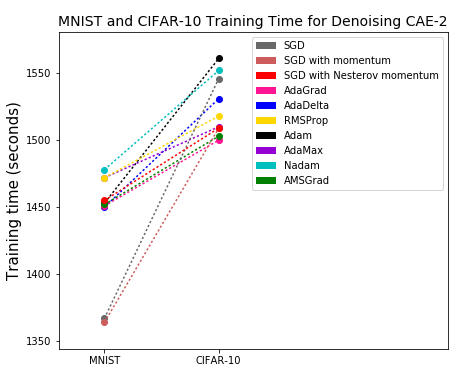}} 
\vspace*{8pt}
\caption{{\it(Top)} MNIST and CIFAR-10 training time for CAE-1 and CAE-2. {\it(Bottom)} MNIST and CIFAR-10 training time for denoising CAE-1 and CAE-2.}
\label{unsupervised_training_time}
\end{figure}  

\section{Conclusion}\label{ch:conclusion}

In this study, the most commonly used optimization algorithms in deep learning are examined. The differences between their working principles are summarized considering pros and cons for each of them. In this context, the importance of adaptive learning algorithms is highlighted. The performances of algorithms are compared on four image datasets empirically. The behaviour of the algorithms during training is observed according to the effects of different image resolutions and neural network architectures. Because of adaptive methods are mostly superior and computationally efficient, their results seem to be better for both tasks. Still, the research to find better adaptive methods continues for deep learning.














\vspace*{-0.01in}
\noindent
\rule{12.6cm}{.1mm}



\biophoto{derya}{{\bf Derya Soydaner} received her
B.Sc., M.Sc. and Ph.D.~degrees in statistics from Mimar Sinan Fine Arts University, Istanbul, Turkey in 2012, 2014 and 2018,\break respectively. She mainly studied on neural networks throughout her entire postgraduate education. Currently, she is a faculty member in the Department of Statistics at Mimar Sinan Fine Arts University. Her research interests include\break pattern recognition, machine learning and image processing.}

\end{document}